\documentclass[runningheads]{llncs}

\usepackage[table, svgnames, dvipsnames]{xcolor}
\usepackage{eccv}

\newcommand{\eat}[1]{}

\usepackage{booktabs}
\usepackage{multirow}
\usepackage{siunitx}
\usepackage{makecell, cellspace, caption}
\usepackage{bm}

\DeclareMathOperator*{\argmax}{arg\,max}

\usepackage{mathtools}
\usepackage{bbm}
\DeclarePairedDelimiterX{\infdivx}[2]{(}{)}{%
  #1\;\delimsize\|\;#2%
}

\usepackage{stmaryrd}

\makeatletter
\newcommand{\xMapsto}[2][]{\ext@arrow 0599{\Mapstofill@}{#1}{#2}}
\def\Mapstofill@{\arrowfill@{\Mapstochar\Relbar}\Relbar\Rightarrow}
\makeatother

\usepackage{enumitem,kantlipsum}
\usepackage{algorithm}

\usepackage{algpseudocode}

\usepackage{mathrsfs}

\algnewcommand{\Initialize}[1]{%
  \State \textbf{Initialize:}
  \Statex \hspace*{\algorithmicindent}\parbox[t]{.8\linewidth}{\raggedright #1}
}
\algnewcommand{\Inputs}[1]{%
  \State \textbf{Inputs:}
  \Statex \hspace*{\algorithmicindent}\parbox[t]{.8\linewidth}{\raggedright #1}
}
\algnewcommand{\Outputs}[1]{%
  \State \textbf{Outputs:}
  \Statex \hspace*{\algorithmicindent}\parbox[t]{.8\linewidth}{\raggedright #1}
}

\definecolor{seal}{HTML}{B711AC}
\definecolor{maroon}{cmyk}{0,0.87,0.68,0.32}

\usepackage{tikz}
\newcommand*\circled[1]{\tikz[baseline=(char.base)]{
            \node[shape=circle,draw,inner sep=1pt] (char) {#1};}}
\usepackage{balance}
\usepackage{eccvabbrv}

\usepackage{graphicx}
\usepackage{booktabs}
\usepackage{multirow}
\usepackage{amsmath}
\usepackage{amssymb}
\usepackage{wrapfig}
\usepackage[accsupp]{axessibility}  

\usepackage{hyperref}

\usepackage{orcidlink}

\begin{document}

\title{Find n' Propagate: Open-Vocabulary 3D Object Detection in Urban Environments} 

\titlerunning{Find n' Propagate}

\author{Djamahl Etchegaray\inst{1}\orcidlink{0009-0000-1154-5389} \and
Zi Huang\inst{1}\orcidlink{0000-0002-9738-4949} \and
Tatsuya Harada\inst{2}\orcidlink{0000-0002-3712-3691} \and Yadan Luo \thanks{Correspondence to Yadan Luo (y.luo@uq.edu.au)}\inst{1}\orcidlink{0000-0001-6272-2971}}

\authorrunning{D.~Etchegaray et al.}

\institute{UQMM Lab, University of Queensland, Brisbane, Australia
\\
\and
The University of Tokyo, Tokyo, Japan. \\ \email{\{uqdetche, helen.huang, y.luo\}@uq.edu.au, harada@mi.t.u-tokyo.ac.jp}}

\maketitle
\vspace{-1ex}

\begin{abstract}
   In this work, we tackle the limitations of current LiDAR-based 3D object detection systems, which are hindered by a restricted class vocabulary and the high costs associated with annotating new object classes. Our exploration of open-vocabulary (OV) learning in urban environments aims to capture novel instances using pre-trained vision-language models (VLMs) with multi-sensor data. We design and benchmark a set of four potential solutions as baselines, categorizing them into either top-down or bottom-up approaches based on their input data strategies. While effective, these methods exhibit certain limitations, such as missing novel objects in 3D box estimation or applying rigorous priors, leading to biases towards objects near the camera or of rectangular geometries. To overcome these limitations, we introduce a universal \textsc{Find n' Propagate} approach for 3D OV tasks, aimed at maximizing the recall of novel objects and propagating this detection capability to more distant areas thereby progressively capturing more. In particular, we utilize a greedy box seeker to search against 3D novel boxes of varying orientations and depth in each generated frustum and ensure the reliability of newly identified boxes by cross alignment and density ranker. Additionally, the inherent bias towards camera-proximal objects is alleviated by the proposed remote simulator, which randomly diversifies pseudo-labeled novel instances in the self-training process, combined with the fusion of base samples in the memory bank. Extensive experiments demonstrate a 53\% improvement in novel recall across diverse OV settings, VLMs, and 3D detectors. Notably, we achieve up to a 3.97-fold increase in Average Precision (AP) for novel object classes. The source code is made available at \href{https://github.com/djamahl99/findnpropagate}{github.com/djamahl99/findnpropagate}.
\end{abstract}

\section{Introduction}\vspace{-1ex}
\label{sec:intro}


LiDAR-based 3D object detection \cite{DBLP:journals/ijcv/MaoSWL23,DBLP:journals/corr/abs-2401-06542,DBLP:conf/iccv/ChenL0BH23} has been well appreciated in recent years owing to its wide applications to self-driving \cite{DBLP:conf/nips/DengQNFZA21,DBLP:conf/eccv/WangLGD20} and robotics \cite{DBLP:conf/iros/AhmedTCMW18, DBLP:conf/iros/MontesLCD20, DBLP:journals/sensors/WangLSLZSQT19}. Despite this, the task aspires to more than it can practically achieve. Established 3D detection baselines \cite{DBLP:conf/cvpr/GeigerLU12, DBLP:conf/corl/HoustonZBYCJOIO20, DBLP:conf/cvpr/SunKDCPTGZCCVHN20, DBLP:conf/nips/MaoNJLCLLY0LYXX21} predominantly focus on a limited vocabulary of classes in urban environments. Key benchmark datasets like KITTI \cite{DBLP:conf/cvpr/GeigerLU12} and Waymo \cite{DBLP:conf/cvpr/SunKDCPTGZCCVHN20} focus their evaluation of detection performance on a mere 3 to 4 common classes, typically including cars, pedestrians and cyclists. Scaling up the range of object classes inevitably entails substantial costs for annotating data with new concepts (\textit{e.g.}, barriers). Consequently, the currently established \begin{wrapfigure}{l}{0.7\linewidth}\vspace{-2ex}
  \begin{center}
\vspace{-2ex}
    \includegraphics[width=\linewidth]{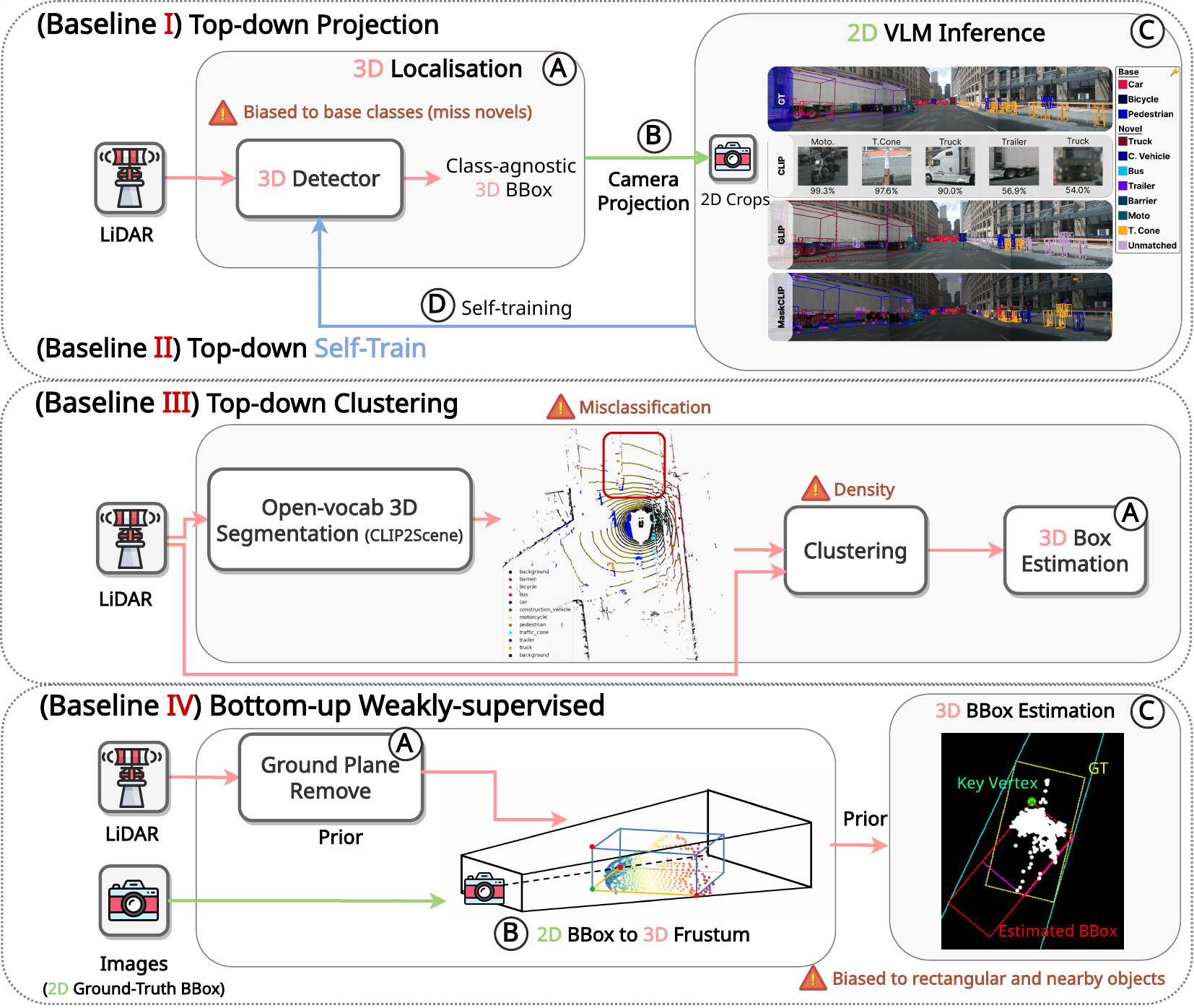}
  \end{center}
\vspace{-4ex}
  \caption{\textbf{The Comparison between four proposed OV-3D Top-down and Bottom-up baselines.}}\label{fig:top-down-bottom-up} \vspace{-6ex}
\end{wrapfigure}methods are not yet applicable to the real-world applications that 3D object detection aspires to solve, as the trained detector cannot identify enough classes to guarantee safe operation. 


\noindent Open-vocabulary (OV) learning presents a forward-thinking approach to recognise new concepts that were absent during training, without the need for labelled data. Existing OV approaches have been mainly applied to 2D object detection: (1) distilling knowledge from large vision-language models (VLMs) such as CLIP \cite{radford2021learning} for feature map matching \cite{gu2021open}, region prompting \cite{zang2022open,wu2023cora}, bipartite matching \cite{DBLP:journals/corr/abs-2310-16667}, and/or (2) employing pseudo-labelled boxes \cite{DBLP:conf/cvpr/ZhongYZLCLZDYLG22, DBLP:conf/eccv/ZhaoZSZKSCM22} or auxiliary grounding data \cite{DBLP:conf/cvpr/LiZZYLZWYZHCG22, DBLP:conf/nips/ZhangZ00LDWYHG22,DBLP:journals/corr/abs-2205-06230, DBLP:journals/corr/abs-2306-09683} as weak supervision in self-training. However, the application of OV learning in LiDAR-based 3D object detection remains \textit{unexplored}, mainly due to the scarcity of VLMs pre-trained on point cloud datasets.

In this work, we investigate the potential of leveraging OV learning for 3D object detection by employing high-resolution LiDAR data (\textsc{Top}) and multi-view imagery (\textsc{Bottom}). As illustrated in Fig. \ref{fig:top-down-bottom-up}, four baseline solutions are designed: (1) \textsc{Top-down Projection}, (2) \textsc{Top-down Self-train}, (3) \textsc{Top-down Clustering}, and (4) \textsc{Bottom-up Weakly-supervised} 3D detection approaches to facilitate novel object discovery in point clouds. The foundation of our \textsc{Top-down} strategies is inspired by advancements in 2D OV learning, where one can regress class-agnostic bounding boxes based on base box annotations and subsequently leverage VLMs for open-vocabulary classification. Based on that, the \textsc{Top-down Self-train} is the variant that further enhances open-vocabulary performance through self-training mechanisms. Beyond mere 2D projections, our third \textsc{Top-down} baseline explores the feasibility of applying open-vocabulary 3D segmentation directly to 3D detection tasks, utilizing clustering techniques for 3D bounding box estimation. Nevertheless, it is observed that \textsc{Top-down} methods can easily overfit to known classes, potentially \textit{overlooking} novel objects with varying sizes and shapes. As shown in the visualisation of Fig. \ref{fig:top-down-bottom-up}, unseen objects that are of vastly different shapes, such as long vehicles like buses or small traffic cones, often go undetected in class-agnostic 3D proposals and are obscured in 2D crops due to occlusion.

The \textsc{Bottom-up} approach presents a cost-effective alternative akin to weakly-supervised 3D object detection, lifting 2D annotations to construct 3D bounding boxes. Different from \textsc{Top-down} counterparts, this approach is training-free and does not rely on any base annotations, potentially making it more generalisable and capable of finding objects with diverse shapes and densities. In Baseline IV, we study FGR \cite{DBLP:conf/icra/WeiSL021} as an exemplar of \textsc{Bottom-up Weakly-supervised} and evaluate its effectiveness in generating novel proposals. FGR starts with removing background points such as the ground plane, then incorporates the human prior into key-vertex localization to refine box regression. However, their study was limited to regressing car objects, as their vertex localization assumes \textit{rectangular} objects which do not hold for other classes (\textit{e.g.}, pedestrians).
To address these limitations, we propose a novel \textbf{\textsc{Find n' Propagate}} approach to maximise the recall rate of novel objects and then propagate the knowledge to distant regions from the camera progressively. We identify most detection failures of novel objects stem from the uncertainties in 3D object orientation and depth. This observation motivates the development of a \circled{2} \textbf{Greedy Box Seeker} strategy that initiates by generating instance frustums for each unique 2D box prediction region, utilizing \circled{1} Region VLMs such as GLIP \cite{DBLP:conf/cvpr/LiZZYLZWYZHCG22}, or pre-trained OV 2D models like OWL-ViT \cite{DBLP:journals/corr/abs-2205-06230}. These frustums are segmented into subspaces across different angles and depth levels to facilitate an exhaustive greedy search for the most apt 3D proposal,  accommodating a wide variety of shapes and sizes. To control the quality of newly generated boxes, we implement a \circled{3} \textbf{Greedy Box Oracle} that employs two key criteria of multi-view alignment and density ranking to select the most probable proposal. The rationale behind that is that 2D predictions predominantly originate from objects near the camera, characterized by dense point clouds and substantial overlap with the 2D box upon re-projection. Recognizing that relying solely on pseudo labels generated from these 2D predictions could bias the detector towards objects near the camera and overlook those that are distant or obscured, we propose a \circled{4} \textbf{Remote Propagator} to mitigate the bias. To augment novel pseudo labels with distant object geometries, geometry and density simulators are employed to perturb pseudo label boxes to farther distances from the camera and mimic sparser structures. The refined 3D proposals are subsequently integrated into a memory bank, facilitating iterative training of the detection model.



\noindent\textbf{Contributions.} This study presents a pioneering endeavour in integrating OV learning with LiDAR-based 3D detection for urban scenarios. We have extensively benchmarked four distinct \textsc{Top-down} and \textsc{Bottom-up} solutions across a range of open-vocabulary protocols, showcasing their versatility and setting a foundation for future studies in this direction. Our approach (1) maximises the recall of novel objects with greedy proposal generation, (2) maintains the precision of proposals with two quality control criteria, (3) introduces copy n' paste and point dropout augmentation strategies, specifically tailored to simulate the geometric characteristics of missed objects in 2D including faraway and sparse objects, thus effectively compensating for the bias inherent in generated proposals from frustums. Empirical results evidence the efficacy of our bottom-up approach, which achieves a remarkable 21\% absolute increase in the recall rate and a 3.9$\times$ enhancement in average precision (AP) for novel objects.

\vspace{-2ex}
\section{Related Work}\label{sec:related}
\vspace{-2ex}


\noindent \textbf{Open-Vocabulary Object Detection (OV-2D)} is the task of training an object detector to recognise concepts beyond the initial training vocabulary. The task was first demonstrated by OVR-CNN \cite{zareian2021open}, which aligned region features with nouns from image-caption pairs. Secondary approaches in OV-2D involve contrastive pre-trained models such as CLIP~\cite{radford2021learning} and SigLIP \cite{DBLP:journals/corr/abs-2303-15343} to establish a broad vocabulary \cite{DBLP:conf/eccv/ZhouGJKM22, gu2021open, DBLP:conf/cvpr/ZhongYZLCLZDYLG22, wu2023cora, zang2022open} from image crops then fine-tune with box annotations for a subset of base classes, and can be divided based on whether they implicitly or explicitly learn about novel instances.  
The first group primarily rely on learning to distil regional knowledge from a pre-trained backbone without losing the open-vocabulary performance. As CLIP was pre-trained on image-level features, a significant gap exists between the input regional features and the pre-trained image ones, leading to poor performance on novel objects. To mitigate the gap, ViLD \cite{gu2021open} trains a Regional Proposal Network (RPN) on top of CLIP, distilling features from the cropped regions to align with the RoIAlign \cite{NIPS2015_14bfa6bb} extracted region features. CORA \cite{wu2023cora} and OV-DETR \cite{zang2022open} use region prompts to align the extracted features with their image-level counterpart. CoDet \cite{DBLP:journals/corr/abs-2310-16667} develop a pretraining method that aligns each region to one noun with bipartite matching. F-VLM\cite{DBLP:conf/iclr/KuoCGPA23} trains detection heads on top of frozen VLMs. CFM-ViT \cite{DBLP:journals/corr/abs-2309-00775} propose a novel pretraining method for reconstructing masked image tokens. The second group explicitly mines novel instances from weak supervision. Some methods use image-caption data for self-training on novel instances. For example, RegionCLIP \cite{DBLP:conf/cvpr/ZhongYZLCLZDYLG22} pretrains using pseudo-boxes generated from extracted nouns, and Detic \cite{DBLP:conf/eccv/ZhouGJKM22} regresses class-agnostic boxes and associates the largest one with the caption. VL-PLM \cite{DBLP:conf/eccv/ZhaoZSZKSCM22} generates pseudo-boxes for novel classes with a class-agnostic RPN and CLIP. Grounding methods directly apply a matching loss to learn from weakly supervised data and fine-tune on detection datasets, including GLIP \cite{DBLP:conf/cvpr/LiZZYLZWYZHCG22, DBLP:conf/nips/ZhangZ00LDWYHG22} and DetCLIP \cite{DBLP:conf/nips/YaoHWLX0LXX22, DBLP:conf/cvpr/YaoHLXZLX23}. They utilise several pre-training datasets and do not remove rare class names. Similarly, OWL-ViT \cite{DBLP:journals/corr/abs-2205-06230, DBLP:journals/corr/abs-2306-09683} finetune on detection and grounding datasets.
Nevertheless, it is non-trivial to produce a direct derivative of 2D works in 3D.

\noindent \textbf{Weakly-supervised 3D Object Detection} is the task of training 3D object detectors from 2D annotations only. FGR \cite{DBLP:conf/icra/WeiSL021}, MTrans \cite{DBLP:conf/eccv/LiuQHQLTW22}, VG-W3D \cite{DBLP:journals/corr/abs-2312-07530} and WM-3D \cite{DBLP:conf/cvpr/TaoHQ0S23} utilise 2D bounding boxes on camera images, whilst WS3D \cite{DBLP:conf/eccv/MengWZSGD20} uses points in BEV view of which FGR, VG-W3D, and WM-3D did not use \textit{any} 3D labels. Methods often design a 3D bounding box estimation process that takes advantage of human \textit{priors} on the shape of the objects of interest, such as vehicles being rectangular. These works have yet to be expanded to larger vocabularies where the shape of the objects is more diverse and perhaps not known during training. 

\noindent \textbf{Open-vocabulary Learning in 3D.} In 3D point-cloud learning \cite{DBLP:conf/iclr/LuoCWYHB23,DBLP:conf/iccv/LuoCF0BH23}, the acquisition of large-scale point-cloud text pairs remains a significant challenge. This limitation has resulted in a scarcity of OV-3D research, particularly in outdoor scenarios predominantly dependent on LiDAR data. A recent study \cite{DBLP:conf/cvpr/LuXWXTKZ23} leverages indoor RGB-D datasets and proposes a top-down approach. Initially, a point-cloud detector is trained to localize unknown objects, of which labels are assigned based on textual prompts. OpenSight \cite{DBLP:journals/corr/abs-2312-08876} provides preliminary results for urban environments but their baselines did not adhere to the open-vocabulary setting as they trained their proposal baseline with full supervision. Triplet cross-modal contrastive learning \cite{xue2023ulip, Peng2023OpenScene, chen2023clip2scene, zeng2023clip2} has been used to integrate different modalities. Nonetheless, this method's applicability is constrained in urban environments. Open-vocabulary learning has also been explored in 3D classification and segmentation. PointCLIP\cite{Zhang_2022_CVPR} was a pioneering effort in utilising CLIP for 3D understanding, optimizing a view adapter in a few-shot setting. Further research \cite{zhu2023pointclip, huang2023clip2point, liu2023partslip} has applied CLIP to point-cloud depth maps for zero-shot classification. Notably, CLIP2Scene \cite{chen2023clip2scene}, CLIP$^2$ \cite{zeng2023clip2}, and Seal \cite{liu2023segment} have used image-to-point correspondence to adapt CLIP for outdoor LiDAR scenes, indicating a growing interest in open-vocabulary learning in complex 3D contexts. Despite these advancements, the core issue in OV-3D remains \textit{unresolved}: the challenge of boosting the recall of unknown objects and effectively learning from noisy and biased proposals. These crucial aspects, which have yet to be thoroughly addressed, will be discussed in detail in Section \ref{sec:method}.


\vspace{-2ex}
\section{Preliminaries}\vspace{-1ex}
\subsection{Task Formulation of OV-3D}\vspace{-1ex}
In the context of OV-3D, we define the base training set as $\mathcal{D}_{\operatorname{B}} = \{\mathfrak{L}, \mathcal{I}, \mathfrak{B}_{\operatorname{B}}\}$, where $\mathfrak{L}$ denotes the collection of raw point clouds, $\mathcal{I}$ the multi-view imagery, and $\mathfrak{B}_{\operatorname{B}}$ corresponds to the set of 3D bounding boxes associated with the base categories $\mathcal{C}^{\operatorname{B}}$. The primary objective of this task is to generalise the 3D detector pretrained on $\mathcal{D}_{\operatorname{B}}$, \textit{i.e.}, $\mathbf{f}(\cdot; \theta): \mathfrak{L}_t\rightarrow \widehat{\mathfrak{B}}_{\operatorname{B}}\cup\widehat{\mathfrak{B}}_{\operatorname{N}}$  so that it can recognise \textit{novel} (subscript $N$) categories $\mathcal{C}^{\operatorname{N}}$ that were not presented in the training corpus in the test sample $\mathfrak{L}_t$, \textit{i.e.}, $\mathcal{C}^{\operatorname{N}}\cap\mathcal{C}^{\operatorname{B}}= \emptyset$.

\vspace{-2ex}
\subsection{Baselines Design}\label{sec:baselines}\vspace{-1ex}
As there is no systematic study of OV-3D in urban scenes, we first design four baselines for benchmarking. The \textsc{Top-down} framework is a straightforward process in which we utilise 3D points to regress a number of box proposals and project them to 2D to acquire labels. It is motivated by the fact that it is inexpensive to acquire open-vocabulary knowledge from images whilst 3D can offer high-precision box proposals. Below, we introduce each method seen in Fig. \ref{fig:top-down-bottom-up}:

\noindent \textbf{Baseline I: \textsc{Top-down Projection}} uses the base bounding boxes $D_{\text{B}}$ to train a detector to predict class-agnostic bounding boxes, which can be projected to 2D to enable the VLM inference. This method can utilise one of many available 2D VLMs, including CLIP \cite{DBLP:conf/icml/RadfordKHRGASAM21}, GLIP \cite{DBLP:conf/cvpr/LiZZYLZWYZHCG22} and MaskCLIP\cite{DBLP:conf/eccv/ZhouLD22}. Because this method uses the base knowledge to train a proposal network it can predict high-precision bounding boxes efficiently but requires costly 3D annotations for base classes. This method may miss many novel objects as it has been trained on the base annotations $\mathcal{D}_{\text{B}}$ and is implicitly biased to avoid novel proposals, reducing its generalisation ability. 


\noindent \textbf{Baseline II: \textsc{Top-down Self-train}}. This baseline introduces a further addition to Baseline I to encourage it to produce more novel object predictions. We encourage the model to predict objects belonging to novel instances with \textit{self-training}. The model will likely make some reasonable novel predictions for objects close to those seen in training, which can be resampled to propagate novel information. However, errors may arise from the VLM's predictions when the objects in cropped regions are occluded or partially observed, which will be accumulated in self-training.


\noindent \textbf{Baseline III. \textsc{Top-down Clustering}.}
We establish the performance of directly clustering points for \textsc{Top-down} proposal generation using density clustering methods such as DBScan \cite{DBLP:conf/kdd/EsterKSX96} and HDBScan \cite{DBLP:conf/pakdd/CampelloMS13} and box fitting from \cite{DBLP:conf/cvpr/LuXWXTKZ23}. To extend the capabilities of the baseline, we establish the performance of an open-vocabulary 3D segmentation model for \textsc{Top-down} proposal generation. CLIP2Scene \cite{chen2023clip2scene} is selected as it is a pioneering work to distil CLIP features in a 3D point cloud encoder for open-vocabulary scene understanding. The proposed Baseline III uses CLIP2Scene to predict the labels of each point, then concatenates the labels and coordinates to be clustered with HDBScan, after which each cluster will be regressed into a 3D box. Issues arise in generating proposals as the zero-shot segmentation model may make misclassifications, and clustering relies on densely distributed objects. 

\noindent \textbf{Baseline IV:  \textsc{Bottom-up Weakly-Supervised}.} To remedy the generalisation problems that arise with \textsc{Top-down} methods, below we detail a \textsc{Bottom-up} alternative. \textsc{Bottom-up} proposal generation is currently explored in the task of weakly-supervised 3D object detection \cite{DBLP:conf/eccv/LiuQHQLTW22, DBLP:journals/corr/abs-2312-07530, DBLP:conf/cvpr/TaoHQ0S23} by using 2D annotations. Among previous works in weakly-supervised 3D object detection, FGR\cite{DBLP:conf/icra/WeiSL021} is the most relevant to our bottom-up definition, regressing 3D boxes from 2D ground-truth and LiDAR \textit{without} any 3D annotations. All weakly-supervised methods apply human prior knowledge to reduce the degrees of freedom and find valid proposals. For FGR, their guiding assumption is that objects are quite rectangular and there exists a vertex with high density in the corresponding right-angle triangle to nearby edges. Further, they assume there exists a flat ground plane where objects lie. These assumptions, however, may impact the model performance when applied to general classes such as non-rectangular pedestrians and cyclists or signs as they are not on the ground.

\begin{figure*}[t]
    \centering\vspace{-1ex}
    \includegraphics[width=1\linewidth]{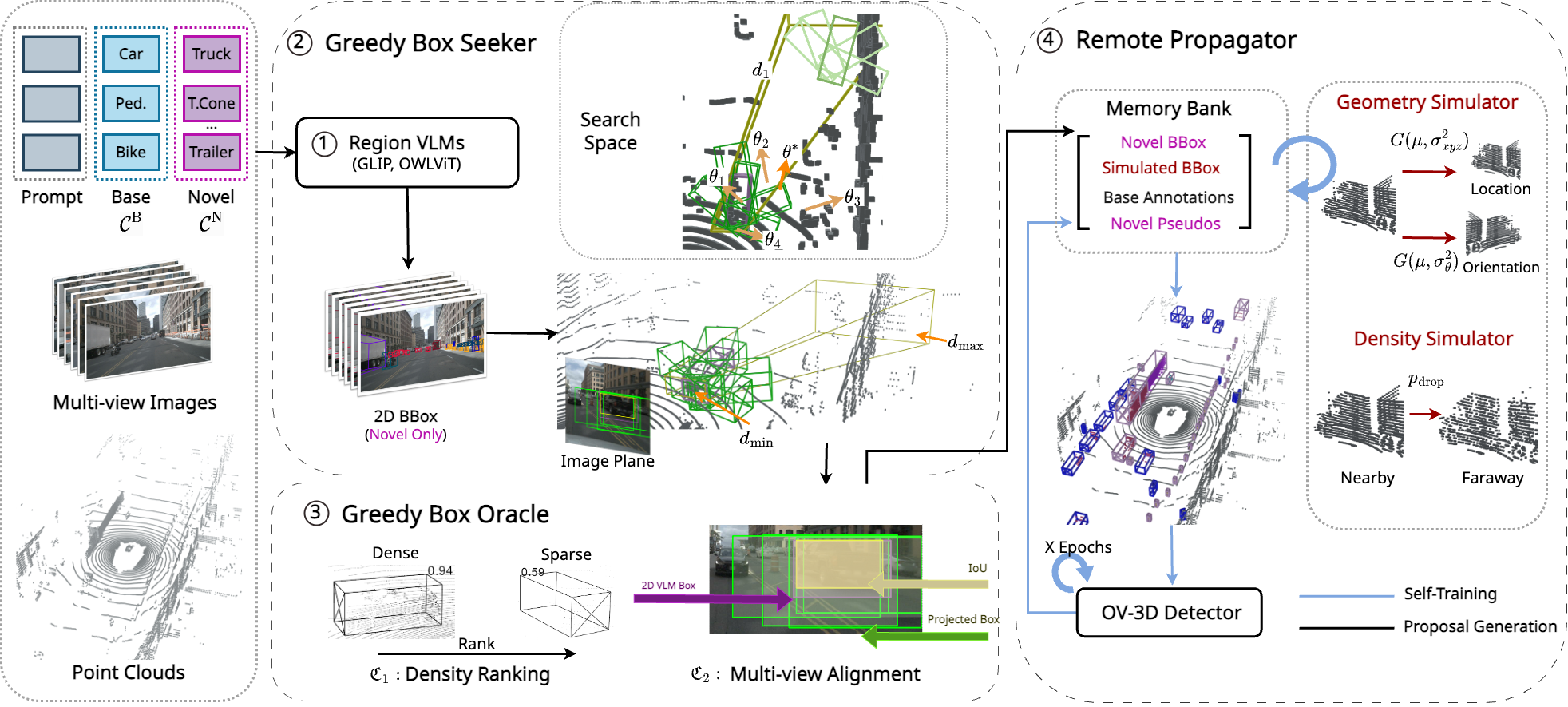}\vspace{-2ex}
    \caption{\textbf{The proposed \textsc{Find n' Propagate} approach.} The framework aims to maximise the recall of novel objects through a Greedy Box Seeker and control the quality of newly identified boxes with Greedy Box Oracle. To propagate the knowledge to distant areas, a Remote Propagator is applied, which allows diverse remote novel instances to be progressively captured.}
    \label{fig:bottom-up}
    \vspace{-3ex}
\end{figure*}
\vspace{-2ex}
\section{Find n' Propagate}\label{sec:method}
\vspace{-1ex}

To avoid the issues from \textsc{Top-down} and \textsc{Bottom-up} OV-3D baselines, we introduce a \textsc{Find n' Propagate} approach to enhance the detection of novel instances in point clouds. The proposed two-stage methodology seeks to initially \circled{1} utilize predictions from region VLMs (\textit{e.g.}, GLIP) or off-the-shelf OV-2D detectors (\textit{e.g.}, OWL-ViT) to enhance the detection of novel objects in proximity to the camera, across a spectrum of orientations and depths, employing a \circled{2} Greedy Box Seeker. To maintain the quality of detections, the newly determined boxes are fed into the \circled{3} Greedy Box Oracle module to filter out low-quality ones. Recognizing that novel objects situated in distant or occluded areas may be overlooked, the method extends to project the predicted 3D proposals through a \circled{4} Remote Propagator, designed to replicate similar geometrical configurations of remote objects. This strategy aims to incrementally improve the capture of novel instances during the self-training process with a coupled memory bank The fundamental components of this approach are illustrated in Fig. \ref{fig:bottom-up}. Below, we first detail the steps of each key module.

\vspace{-3ex}
\subsection{Greedy Box Seeker}\vspace{-1ex}
To identify novel classes $\mathcal{C}^N$ in \circled{1}, GLIP or OWL-ViT is employed to generate bounding boxes $\{\mathfrak{B}_{\operatorname{2D}}^m\}_{m=1}^M$ across all $M$ camera views. Subsequently, these 2D bounding boxes are lifted into 3D space, by creating \textit{frustums} \cite{DBLP:conf/cvpr/QiLWSG18} that define a 3D search space to help localise the amodal 3D bounding box, as shown in Fig. \ref{fig:bottom-up}. As a frustum may capture a wide spread of points including foreground objects and background ones, below we refine the search space. 

\noindent \textbf{Box Search Space.}
Given 2D bounding box $\mathfrak{B}_{\operatorname{2D}}^m$, we define its corresponding frustum $\mathfrak{U^m}$, with depth values $d$ in the range $[0, \infty)$. To refine the search space for 3D proposal $\mathfrak{B}_{\operatorname{3D}}$, we use the lower and upper \textit{quantiles}, $d_{\text{min}}$ and $d_{\text{max}}$ as depth bounds. The resultant search space $\mathbb{R}_{\mathfrak{U^m}}$ in each frustum is the Cartesian product of the boundary sets:\vspace{-1ex}
\begin{equation*}
    \mathbb{R}_{\mathfrak{U^m}} := \{ u_{\text{min}}, u_{\text{max}}\} \times \{ v_{\text{min}}, v_{\text{max}}\} \times \{ d_{\text{min}}, d_{\text{max}} \},\vspace{-1ex}
\end{equation*} where $(u_{\text{min}}, u_{\text{max}})$, $(v_{\text{min}}, v_{\text{max}})$ are 2D top-left and bottom-right coordinates of $\mathfrak{B}_{\operatorname{2D}}^m$. For each novel class $c\in\mathcal{C}^{\operatorname{N}}$, we consider a pre-defined anchor box $\mathcal{A} = (w, l, h) * \gamma$, where $w, l, h$ represents the pre-defined width, length and heights for the novel class $c\in\mathcal{C}^{\text{N}}$ and $\gamma$ is a scaling factor that controls the variety of anchor box size, respectively. The anchor boxes are taken from OpenPCDet. $\gamma$ is empirically sampled from the range of $(0.95, 1.2)$. As illustrated in Fig. \ref{fig:bottom-up}, we divide each frustum by considering $k_{\text{d}}$, $k_{\text{s}}$ and $k_{\text{o}}$ intervals in the depth $d$, scale $\gamma$ and orientation $\theta$ axes, where we greedily search the optimal placement of 3D proposals by considering $k_{\text{o}}$ different poses of the anchor box $\mathcal{A}$ within the range of $(0, \pi)$. To this end, this process results in a pool of 3D proposal candidates from each frustum $\mathcal{D}_{\text{f}} = \{\bar{\mathcal{A}}_i\}_{i=1}^{k_{\text{d}}\times k_{\text{o}}\times k_{\text{s}}}$, where $\bar{\mathcal{A}}_i=(c_x, c_y, c_z, w,l,h, \theta)$ includes the anchor size, centre location and orientation. Considering there are a large number of noisy proposal candidates, we introduce the following \textit{quality control} step that ranks each candidate based on two key metrics to select the most accurate representation of the target 3D object.

\vspace{-3ex}
\subsection{Greedy Box Oracle}\vspace{-1ex}
Proposals from the Greedy Box Seeker will contain many background and poorly regressed boxes we reduce the number of proposals by selecting the best candidate for each frustum. Given a set of proposals $\mathcal{D}_{\text{f}}$ obtained from Greedy Box Seeker, we produce a composite score $\mathfrak{C}$ based on the sum of two normalised criteria, as shown in Fig. \ref{fig:bottom-up}. The score is used to rank the proposals to find the best one for each frustum. The score measures the supporting evidence for each box in both 2D and 3D domains, using Criterion 1 and Criterion 2 respectively. Criteria 1 reflects the positive attribution of points to the box, and Criteria 2 the alignment with the 2D bounding box.


\noindent \textbf{Criterion 1: Density Ranking}. 
Central to our evaluation is the point density within each proposal. Given that camera frustums capture a mix of object-related and extraneous points, distinguishing between the foreground and background is crucial. Our approach introduces a bias toward proposals with high point density. In particular, for LiDAR points, $\mathfrak{L} =\{ (x_i, y_i, z_i)\}_{i=1}^N$ and proposals $\{\bar{\mathcal{A}}^{m}\}_{m=1}^M$, we transform their coordinates to each proposals local coordinates $(\hat{x}_i^m, \hat{y}_i^m, \hat{z}_i^m)$:\vspace{-2ex}
\begin{align}\nonumber
    \hspace{-1ex}\begin{bmatrix}
        \hat{x}_i^m \\ \hat{y}_i^m \\ \hat{z}_i^m \\ 1
    \end{bmatrix}=\begin{bmatrix}\cos{(-\theta_m)} & \sin{(-\theta_m)} & 0 &  - c_x^m \\ -\sin{(-\theta_m)} & \cos{(-\theta_m)} & 0 &  - c_y^m \\ 0 & 0 & 1 &  - c_z^m \\ 0 & 0 & 0 & 1\end{bmatrix}
    \begin{bmatrix}
        x_i \\ y_i \\ z_i \\ 1
    \end{bmatrix},\label{eqn:coord_trans}
\end{align}
where the middle translation matrix transforms global LiDAR coordinates to proposal relative coordinates. To calculate the point density of each proposal, we tally the number of points falling within each proposal and subsequently rank them by dividing this count by the maximum density across all proposals:\vspace{-1ex}
\begin{equation}\label{eq:c1}\hspace{-2ex}
    \mathfrak{C}_2(\mathfrak{L},\bar{\mathcal{A}}^{m}) = \frac{\sum_{i = 1}^N \mathbbm{1}_{\text{inBox}} \left([\hat{x}_i^k, \hat{y}_i^k, \hat{z}_i^k], \, \bar{\mathcal{A}}^{m}\right)}
    {\text{max}_{j\in[M]} \sum_{i = 1}^N \mathbbm{1}_{\text{inBox}} ([\hat{x}_i^j, \hat{y}_i^j, \hat{z}_i^j ], \, \bar{\mathcal{A}}^j)},
\end{equation}
\begin{equation}\nonumber
    \mathbbm{1}_{\text{inBox}}([\hat{x}_i^k, \hat{y}_i^k, \hat{z}_i^k], \bar{\mathcal{A}}^{m}) = \begin{cases}
    1 & \left|[\hat{x}_i^k, \hat{y}_i^k, \hat{z}_i^k]\right| \leq   [\frac{w_k}{2},\frac{l_k}{2}, \frac{h_k}{2}] \\ 
    0 & \text{otherwise}.
    \end{cases}
\end{equation}
This criterion assigns higher rankings to densely packed proposals over sparser ones, which are more likely to represent the object of interest. However, it's worth noting that this criterion \textit{alone} may not be sufficient, as there's a possibility that the background could exhibit denser points.

\noindent \textbf{Criterion 2: Multi-view Alignment}. Dense areas do not imply strong orientation bias, as boxes should maintain most of their density with orientation error. We assume that 2D instance boxes should \textit{tightly bound} the object of interest in the image plane, aiming to maintain the tight bound with our proposals. To evaluate the suitability of a proposal and its associated GLIP bounding box, we introduce the image crop IoU criterion that involves projecting the proposal onto the camera view and calculating the IoU with 2D predictions: \eat{We reject regions with low image crop IoU, and then rank the rest, where the score is derived from all criteria.}
\vspace{-1ex}
\begin{equation}\label{eq:c2}
   \mathfrak{C}_2(\bar{\mathcal{A}}^{m}) = \texttt{IoU}(\mathbf{P}(\bar{\mathcal{A}}^{m}), \mathbf{S}^m).\vspace{-1ex}
\end{equation}
By combing Eq. \eqref{eq:c1} and  Eq. \eqref{eq:c2}, the filtered proposal candidates $\mathcal{D}_{\text{f}}^*$ are obtained by ranking:\vspace{-2ex}
\begin{equation}\nonumber
    \mathcal{D}_{\text{f}}^* = \argmax_{\mathcal{A}^{m}\in\mathcal{D}_{\text{f}}} \mathfrak{C}, \mathfrak {C} = \mathfrak{C}_1(\mathfrak{L}, \bar{\mathcal{A}}^{m}) + \alpha_{\operatorname{IoU}}\mathfrak{C}_2(\bar{\mathcal{A}}^{m}),
    \vspace{-2ex}
\end{equation}
where $\alpha_{\operatorname{IoU}}$ is the coefficient for the alignment criterion. 


\vspace{-2ex}
\subsection{Remote Propagator}\label{sec:self-train}\vspace{-1ex}
The Greedy Box Seeker and Oracle will produce proposals based on the assumption that the objects are close to the camera and characterised by dense and structured point clouds. This can, however, \textit{limit} the 3D detector's ability to identify objects in sparser point clouds or varying contexts. To address this limitation and counteract biases that may accumulate during self-training, we introduce \textbf{Remote Progagator} to propagate knowledge to faraway regions. 

\noindent \textbf{Memory Bank $\mathcal{Q}$.} To start with, we construct a memory bank $\mathcal{Q}$ to support the multi-round near-to-far self-training. As illustrated in Fig. \ref{fig:bottom-up}, $\mathcal{Q}$ takes four data sources including base GT $\mathcal{D}_{\text{B}}$, searched boxes from Oracle $\mathcal{D}^*_{\text{f}}$, simulated remote objects and high-confidence pseudo-labelled boxes $\mathcal{D}_{\text{PL}}$ after the detector get re-trained. We apply filtering strategies to ensure data quality and \textit{consistency} across these varied sources, with further details available in the supplementary material. Below, we provide the details of two proposed geometry and density simulators, which randomly copy and diversify boxes to mimic distant objects.





\noindent \textbf{Geometry Simulator.} 
Different from standard GT sampling, which merely collects samples from an offline dataset and integrates them into new scenes $\mathfrak{L}$ with \textit{static} positions and orientations, we introduce an innovative strategy to augment the object geometry characteristics for both proposals from Oracle $\mathcal{D}^*_{\text{f}}$ and pseudo-labelled novel 3D boxes $\mathcal{D}_{\text{PL}}$. To ensure only high-quality objects are utilized for augmentation, we maintain a fixed-size class-wise queue in $\mathcal{Q}$ to comprise high-confidence objects. This queue is updated when capacity permits or when newly acquired pseudo-labels present objects of superior confidence. This geometry simulator translates LiDAR points into the reference frame of each collected box, retaining points within the box boundaries as determined by the function $\mathbbm{1}_{\text{inBox}}(\cdot; \cdot)$. During training, we randomly select $N_{\text{paste}}$ novel class labels for each iteration, followed by random sampling of indices $k_1, \ldots, k_{N_{\text{paste}}}$ from the range $[1, |\mathcal{Q}|]$.  For each sampled box $\mathfrak{B}_k$, Gaussian noises $G(\mu_{\theta}, \sigma^2_{\theta})$ and $G(\mu_{xyz}, \sigma^2_{xyz})$ are introduced to its geometric locations and orientation angles with standard deviations $\sigma_{\text{xyz}}$ and $\sigma_{\theta}$. 
 Collision avoidance in the target scene is ensured by rejecting any overlapping objects, both among themselves and with pre-existing scene objects, and we persist in the search for suitable placement locations. This derived strategy helps improve the model's generalisation to novel objects that were previously undetected by the Oracle or pseudo-labelled proposals, while also aiding in the precise estimation of object orientations.

\noindent \textbf{Density Simulator.}  To further address the limitation where Oracle proposals predominantly encompass near objects with high point density, we introduce the density simulator to mimic a sparser point cloud structure: we randomly eliminate a fraction of the points from these dense point cloud samples. When we sample an instance from the queue of collected objects in the memory bank, we decide to drop out points with probability $p_{\text{drop}}$ and then randomly drop an amount N$_{\text{drop}} \sim \mathbb{U}\left[0, \frac{N}{2}\right]$.  This simulation mimics the naturally sparse point distribution observed in distant or occluded objects in urban scenes thereby enhancing the detector's robustness.





\vspace{-2ex}
\section{Experiments}\vspace{-2ex}
    
    \subsection{Experimental Set-up}\vspace{-1ex}
    \noindent \textbf{Dataset.} To verify the effectiveness of the proposed OV-3D approaches, extensive experiments are conducted on  nuScenes\cite{DBLP:conf/cvpr/CaesarBLVLXKPBB20} and KITTI \cite{DBLP:conf/cvpr/GeigerLU12}. nuScenes contains 1000 driving sequences, with 700, 150, 150 sequences for training, validation, and testing, respectively. Each 20-second sequence has keyframes annotated at 2Hz, totalling approximately 1.4 million boxes. The dataset includes RGB images from six cameras covering a 360-degree horizontal FOV, with views from the front, front left, front right, back, back left, and back right, all at 1600×900 resolution. Ten object classes are annotated: car, truck, bus, trailer, construction vehicle, pedestrian, motorcycle, bicycle, barrier, and traffic cone. KITTI consists of 3,712 training samples (\textit{i.e.,} point clouds) and 3,769 \textit{val} samples. The dataset includes a total of 80,256 labelled objects with three commonly used classes for autonomous driving: cars, pedestrians, and cyclists. 
    
    \noindent \textbf{Open-vocabulary Settings.} To study open-vocabulary performance, we have developed two protocols, varying in complexity from moderate to challenging.\vspace{-1ex}
    \begin{itemize}[leftmargin=*]
        \item \textcolor{seal}{\S\textsc{Setting 1}:} $|\mathcal{C}^{\operatorname{B}}|$=6, $|\mathcal{C}^{\operatorname{N}}|$=4. We categorize six classes as base classes: car, construction vehicles, trailer, barrier, bicycle, and pedestrian while treating the remaining four as novel categories. This grouping is motivated by \cite{DBLP:conf/cvpr/YinZK21}, which pairs novel and known classes based on size similarities (\textit{e.g.}, trailer vs. bus, bicycle vs. motorcycle). Implementing this strategy facilitates a balanced recall rate for detecting novel classes, ensuring their identification is on par with base categories.
        \item\textcolor{seal}{ \S\textsc{Setting 2}:} $|\mathcal{C}^{\operatorname{B}}|$=3, $|\mathcal{C}^{\operatorname{N}}|$=7. We further delve into a more challenging scenario by only having three common classes, \textit{i.e.}, car, pedestrian, and bicycle as bases following KITTI \cite{DBLP:conf/cvpr/GeigerLU12}.  This approach aligns more closely with real-world applications, acknowledging the reality that the accurate detection and regression of novel objects cannot always be assured. 
        \item\textcolor{seal}{\S\textsc{Setting 3}:} $|\mathcal{C}^{\operatorname{N}}|$=10. In our third and more ambitious test setting, we evaluate proposal methods with no base classes, leaving all ten classes as novels. 
    \end{itemize}

    \noindent \textbf{Evaluation Metrics.}
    For evaluating 3D object detection, two key metrics are utilised: mean Average Precision (mAP) and the nuScenes detection score (NDS). The NDS metric is a composite score, calculated as a weighted average of mAP along with various attribute metrics \textit{e.g.}, translation, scale, orientation, velocity, and other box attributes. We further introduce $\text{AP}_{\text{B}}$ and $\text{AP}_{\text{N}}$, which distinctly calculates the mean average precision across the base classes and novel classes, respectively. Average recall (AR$_\text{N}$) is calculated for novel classes. We evaluate the performance of learning methods with Transfusion \cite{DBLP:conf/cvpr/BaiHZHCFT22} and Centerpoint \cite{DBLP:conf/cvpr/YinZK21} as they are different architectures with different label assignment methods that both achieve strong performance on nuScenes. Implementation details and \texttt{source code} are at \href{https://github.com/djamahl99/findnpropagate}{github.com/djamahl99/findnpropagate}.

\vspace{-2ex}
\subsection{Main Results} \vspace{-1ex}
\begin{table*}[t]\caption{\textbf{Open-vocabulary evaluation on the nuScenes dataset with four novel classes (\textcolor{seal}{$\S$ \textsc{Setting} 1}).} $^*$ indicates the result with class-agnostic Transfusion and -f denote the variants with logit fusion. All scores are given in percentage. ``C2S'' denotes CLIP2Scene, ``Trans'' as Transfusion, and ``Center'' as Centerpoint.\vspace{-2ex}}
    \centering
    \resizebox{1\linewidth}{!}{
    \begin{tabular}{lccc | S[table-format=2.2]S[table-format=2.2]S[table-format=2.2]S[table-format=2.2]S[table-format=2.2]S[table-format=2.2] |S[table-format=2.2]S[table-format=2.2]S[table-format=2.2]S[table-format=2.2]| S[table-format=2.2]S[table-format=2.2]S[table-format=2.2]S[table-format=2.2] c}
    \toprule
    &  & & & \multicolumn{6}{c}{\textsc{Base}}      & \multicolumn{4}{c}{\textcolor{seal}{\textsc{Novel}}}     & \multicolumn{5}{c}{\textsc{Overall}}     \\
Method & Box$_{\text{3D}}$ & VLM   & Arch. & \multicolumn{1}{c}{Car} &\multicolumn{1}{c}{Const.} &\multicolumn{1}{c}{Trai.} & \multicolumn{1}{c}{Barr.} & \multicolumn{1}{c}{Bic.} & \multicolumn{1}{c}{Ped.} &\multicolumn{1}{c}{\textcolor{seal}{Truck}} & \multicolumn{1}{c}{\textcolor{seal}{Bus}} & \multicolumn{1}{c}{\textcolor{seal}{Motor.}} & \multicolumn{1}{c}{\textcolor{seal}{Cone.}} & \multicolumn{1}{c}{mAP} & \multicolumn{1}{c}{NDS} & \multicolumn{1}{c}{AP$_{\text{B}}$} & \multicolumn{1}{c}{~\textcolor{seal}{AP$_{\text{N}}$}} &  \multicolumn{1}{c}{~\textcolor{seal}{AR$_{\text{N}}$}}   \\
\midrule %
 \multirow{16}{*}{\rotatebox[origin=c]{0}{\textsc{Top-Projection}}}
 & Base   & - & Trans. & 86.53 &26.17  &42.21    &68.46    &52.73    & 73.56        & 0.00    & 0.00  & 0.00  & 0.00   & 34.97 & 40.02 & \textbf{58.28}      & 0.00  & -    \\
  & Base   & - & Center. & 84.08 & 16.62  & 35.59 & 67.33 & 39.53 & 85.25 & 0.00 & 0.00 & 0.00 & 0.00 & 32.84 & 38.75 & 54.74 & 0.00 & -\\
 & Base & CLIP$^*$ & Trans.    & 54.89 & 0.00    & 0.00      & 1.48      & 22.92     & 28.64        & 3.09    & 4.95  & 4.33         & 0.53   & 12.08 & 32.98 & 17.99       & 3.22    & -\\
 & Base &  & Trans.     & 55.82 & 0.00    & 0.00      & 1.16      & 24.52     & 27.55        & 2.96    & 4.30  & 4.62         & 0.38   & 12.13 & 32.89 & 18.18       & 3.06    & - \\
& Base   & \multirow{-2}{*}{\rotatebox[origin=c]{0}{CLIP}} & Center. & 57.39 & 0.00   & 0.00 & 0.00 & 30.24 & 29.80 & 2.96 & 2.45 & 6.04 & 0.00 & 12.89 & 29.35 & 19.57 & 2.86  & -\\
 & Base &  & Trans.   & 83.09 & 4.18    & 34.58     & 61.87     & 49.30     & 67.28        & 0.00    & 5.54  & 11.14        & 0.00   & 31.70 & 43.24 & 50.05       & 4.17    & -\\
 & Base   & \multirow{-2}{*}{\rotatebox[origin=c]{0}{CLIP-f}} & Center. & 66.49 & 2.21   & 17.89 & 44.27 & 40.60 & 65.88 & 0.30 & 0.00 & 0.20 & 0.00 & 23.78 & 42.98 & 39.56 & 0.13 & -\\
& Base & & Trans. &56.93  &0.10  &0.00  &3.55  &30.93  &32.86  &2.40  &0.00  &0.40  &0.00  &12.72  &33.30  &20.73  &0.70 & -\\
& Base &\multirow{-2}{*}{\rotatebox[origin=c]{0}{Mask}} & Center. & 50.13 & 0.13   & 0.00 & 3.61 & 31.43 & 40.93 & 1.20 & 0.00 & 0.00 & 0.00 & 12.74 & 33.51 & 21.04 & 0.30 & -\\
 & Base& & Trans. &71.66  &2.58  &20.63  &43.62  &41.40  &54.51  &1.23  &0.46  &3.48  &0.00  &23.96  &42.77  &39.07  &1.29  & - \\
  & Base&\multirow{-2}{*}{\rotatebox[origin=c]{0}{Mask-f}} & Center. & 66.50 & 2.22   & 17.89 & 44.22 & 40.60 & 65.88 & 0.30 & 0.00 & 0.19 & 0.00 & 23.78 & 42.98 & 39.55 & 0.12 & -\\
& Base& & Trans.   & 69.39 & 2.42    &2.88 &43.88 &43.00 &62.57 &3.60 &0.00 &12.56 &1.97 &24.23 &42.71 &37.36 &4.53 & -\\
 & Base   & \multirow{-2}{*}{\rotatebox[origin=c]{0}{GLIP}} & Center. & 65.20 & 2.07   & 3.44 & 43.85 & 44.68 & 71.36 & 2.34 & 0.00 & 3.27 & 1.22 & 23.74 & 42.88 & 38.43 & 1.71 & -\\
& Base&  & Trans.   & 78.71 & 6.95    & 11.74     & 63.88     & 44.88     & 73.22        & 7.35    & 2.57  & 17.15        & 0.00   & 30.64 & 46.40 & 46.56       & 6.77 & -\\  
 & Base   & \multirow{-2}{*}{\rotatebox[origin=c]{0}{GLIP-f}} & Center. & 76.91 & 4.59   & 10.78 & 63.15 & 39.00 & 84.88 & 5.15 & 0.12 & 4.95 & 0.00 & 28.95 & 46.07 & 46.55 & 2.55 & -\\
 (Upper Bound)& \textbf{GT}$_{\text{3D}}$    & CLIP & Trans. & 53.10 &	9.47 & 0.00	& 0.72	& 39.95 & 31.20 & 16.64 & 62.65	& 37.10 & 34.05 & 29.14 & 45.59 & 23.48 & 37.61    & - \\ 
 \midrule
  \multirow{2}{*}{\rotatebox[origin=c]{0}{\textsc{Top-Clustering}}} & HDBScan & C2S\cite{chen2023clip2scene}  &-  & 4.01 & 0.00 & 0.00 & 0.00 & 0.00 & 0.00 & 1.38 & 0.37 & 0.00 & 1.27 & 0.70 & 6.10 & 0.67 & 0.76 & 7.83      \\ 
  & DBScan & C2S\cite{chen2023clip2scene}  &-  & 3.07 & 0.00 & 0.00 & 0.00 & 0.00 & 0.00 & 1.86 & 0.72 & 0.00 & 0.00 & 0.57 & 3.33 & 0.51 & 0.65 & 7.76     \\ 
   
  \midrule
 \midrule
  & Base+ST & CLIP    & Trans. &  85.37 & 18.64 & 32.37 & 68.59 & 35.08 & 85.18 & 12.49 & 4.80 & 8.50 & 0.11 & 35.11 & 46.34 & 54.20 & 6.47 & 39.40 \\
  \multirow{-2}{*}{\rotatebox[origin=c]{0}{\textsc{Top-SelfTrain}}} & Base+ST & GLIP & Trans.  & 85.50 & 17.81 & 32.67 & 68.04 & 33.54 & 85.10 & 12.82 & 0.01 & 2.34 & 0.00 & 33.78 & 41.86 & 53.78 & 3.79 & 38.68 \\
 \midrule
    
  \rowcolor{Gainsboro!60} &Seeker+ST & OWL & Center.  & 78.70 & 11.71 & 26.09 & 64.11 & 31.44 & 81.59 & 0.80 & 14.18 & 20.80 & 21.23 & 35.06 & 40.85 & 48.94 & 14.25 & 28.39 \\
  \rowcolor{Gainsboro!60} &Seeker+ST & GLIP & Center.  & 78.81 & 13.14 & 26.48 & 63.39 & 33.40 & 81.99 & 7.07 & 19.10 & 9.51 & 40.92 & 37.38 & 39.82 & 49.54 & 19.15 & 41.12 \\
 \rowcolor{Gainsboro!60} &Seeker+ST  & OWL & Trans.    & 84.98 & 17.35 & 31.93 & 65.63 & 34.70 & 83.94 & 20.76 & 26.52 & \textbf{34.76} & 24.60 & 42.52 & 45.13 & 53.09 & 26.66 & \textbf{60.10}  \\
  \rowcolor{Gainsboro!60} \multirow{-4}{*}{\textbf{\textsc{Find n' Propagate}}} &Seeker+ST &GLIP & Trans.  & 84.30 & 15.91 & 30.38 & 69.26 & 31.48 & 83.55 & \textbf{26.17} & \textbf{28.43} & 34.18 & \textbf{45.83} & \textbf{44.95} & \textbf{47.87} & 52.48 & \textbf{33.65} & 58.46  \\
 \bottomrule
    \end{tabular}
}
    \vspace{-2ex}
    \label{tab:6-class-all}
\end{table*}

\begin{table*}[t]\caption{\textbf{Evaluation on nuScenes with 10 novel classes (\textcolor{seal}{$\S$ \textsc{Setting} 3}).}\vspace{-2ex}}
    \centering
    \resizebox{1\linewidth}{!}{
    \begin{tabular}{lccc | S[table-format=2.2]S[table-format=2.2]S[table-format=2.2]S[table-format=2.2]S[table-format=2.2]S[table-format=2.2] S[table-format=2.2]S[table-format=2.2]S[table-format=2.2]S[table-format=2.2]| S[table-format=2.2]S[table-format=2.2]}
    \toprule
    &  & & &  \multicolumn{10}{c}{\textcolor{seal}{\textsc{Novel}}}     & \multicolumn{2}{c}{\textsc{Overall}}     \\
Method & Box$_{\text{3D}}$ & VLM   & Arch. & \multicolumn{1}{c}{\textcolor{seal}{Car}} &\multicolumn{1}{c}{\textcolor{seal}{Const.}} &\multicolumn{1}{c}{\textcolor{seal}{Trai.}} & \multicolumn{1}{c}{\textcolor{seal}{Barr.}} & \multicolumn{1}{c}{\textcolor{seal}{Bic.}} & \multicolumn{1}{c}{\textcolor{seal}{Ped.}} &\multicolumn{1}{c}{\textcolor{seal}{Truck}} & \multicolumn{1}{c}{\textcolor{seal}{Bus}} & \multicolumn{1}{c}{\textcolor{seal}{Motor.}} & \multicolumn{1}{c}{\textcolor{seal}{Cone.}} & \multicolumn{1}{c}{\textcolor{seal}{AP$_{\text{N}}$}} & \multicolumn{1}{c}{NDS}   \\
\midrule %
    \multirow{1}{*}{\textsc{Top-Clustering}} & DBScan & GLIP &- & 2.70 & 0.96 & 0.00 & 0.96 & 0.00 & 0.00 & 1.33 & 0.62 & 0.24 & 0.00 & 0.68 & 4.63 \\
   & DBScan & \textbf{GT}$_{\text{2D}}$ &- &  2.37 & 0.94 & 0.00 & 7.80 & 0.90 & 0.45 & 1.40 & 0.54 & 0.79 & 0.00 & 1.52 & 7.09 \\ 
  \midrule
     & & OWL & - & 14.20 & 1.80 & 0.00 & 0.10 & 5.20 & 19.40 & 2.50 & 4.00 & 5.50 & 1.10 & 5.40 & 12.40 \\
    \multirow{-2}{*}{OpenSight \cite{DBLP:journals/corr/abs-2312-08876}} & & Detic & - & 15.10 & 2.10 & 0.00 & 0.10 & 6.20 & 21.10 & 2.90 & 4.20 & 6.10 & 0.80 & 5.80 & 12.70 \\
  \midrule
     \rowcolor{Gainsboro!60}  &Seeker & OWL & - & \textbf{25.14} & 0.76 & 0.00 & 0.30 & 19.21 & 12.24 & 6.07 & 5.89 & 18.01 & 18.89 & 10.65 & 18.30 \\ 
    
   \rowcolor{Gainsboro!60}\multirow{-2}{*}{\textsc{Ours}}  &Seeker & GLIP & - &  24.28 & \textbf{4.14} & \textbf{0.15} & \textbf{4.39} & \textbf{34.65} & \textbf{22.80} & \textbf{8.58} & \textbf{11.09} & \textbf{35.82} & \textbf{21.26} & \textbf{16.72} & \textbf{22.40} \\ 
    \rowcolor{Gainsboro!60} (Upper Bound) &Seeker & \textbf{GT}$_{\text{2D}}$ & - &  16.09 & 3.26 & 0.37 & 34.28 & 38.22 & 39.03 & 7.41 & 5.54 & 30.33 & 56.45 & 23.10 & 22.83 \\
 \bottomrule
    \end{tabular}
}
    \vspace{-3ex}
    \label{tab:0-class-all}
\end{table*}

\begin{table}[t]
    \centering
    \caption{\textbf{Comparisons under the challenging \textcolor{seal}{$\S$ \textsc{Setting} 2}.}\vspace{-2ex}}
    \label{tab:3-class}\vspace{-1ex}
    \resizebox{0.7\linewidth}{!}{
    \begin{tabular}{l lccccc}
    \toprule
      \multicolumn{1}{l}{Method} &VLM & Architecture & \multicolumn{1}{c}{mAP} & \multicolumn{1}{c}{NDS} & \multicolumn{1}{c}{AP$_\text{B}$} & \multicolumn{1}{c}{\textcolor{seal}{AP$_\text{N}$}} \\
    \midrule
    \textsc{Base} &- & Transfusion & 21.85   & 24.42   & 72.85    & 0.00   \\
    \textsc{Base} &- & Centerpoint  & 21.67&	24.40 &	72.23 &	0.00  \\
    \textsc{Top-down Proj.} &CLIP-f  & Transfusion &   21.86   & 24.42   &\textbf{ 72.86}    & 0.00   \\
    \textsc{Top-down Proj.} &CLIP-f  & Centerpoint &  18.13 &	33.07 &	57.37 &	1.31     \\
    \textsc{Top-down Proj.} &GLIP-f  &  Transfusion & 21.66 & 30.49   & 71.36 & 0.36   \\
    \textsc{Top-down Proj.} &GLIP-f  &  Centerpoint & 22.71 &	33.67 &	69.78 &	2.54   \\
    \textsc{Top-down Proj.} &MaskCLIP-f   &  Transfusion & 23.78   & \textbf{42.98}   & 62.46    & 2.23   \\
    \textsc{Top-down Proj.} &MaskCLIP-f   &  Centerpoint &  19.28 &	33.17 &	60.00 &	1.83   \\
     \rowcolor{Gainsboro!60}\textsc{Find n' Propagate}~    &GLIP &  Transfusion &  31.44 & 34.53 & 67.41 & \textbf{16.03}   \\ 
     \rowcolor{Gainsboro!60}\textsc{Find n' Propagate}~    &GLIP &  Centerpoint & \textbf{37.38} & 40.28 & 49.99 & \textbf{18.46} \\
    \bottomrule
    \end{tabular}}\vspace{-3ex}
\end{table}

\begin{table}[t]
\centering
\caption{\textbf{\textsc{Bottom-up} proposals evaluation on KITTI validation set.} All values in \%. GT$_{\text{2D}}$ is used for all methods. Below are classes that were not tested in FGR.\vspace{-2ex}}
\resizebox{0.7\linewidth}{!}{
\label{tab:kitti_proposal}
\begin{tabular}{l  S[table-format=2.2]S[table-format=2.2]S[table-format=2.2] S[table-format=2.2]S[table-format=2.2]S[table-format=2.2]}
\toprule
   & \multicolumn{3}{c}{\textcolor{seal}{Cyclist AP$_{\text{3D}}$}} & \multicolumn{3}{c}{\textcolor{seal}{Pedestrian AP$_{\text{3D}}$}}  \\

  \multicolumn{1}{l}{Method} & \multicolumn{1}{l}{Easy} & \multicolumn{1}{l}{Moderate} & \multicolumn{1}{l}{Hard} & \multicolumn{1}{l}{Easy} & \multicolumn{1}{l}{Moderate} & \multicolumn{1}{l}{Hard} \\
\midrule %
   \textsc{Fgr}\cite{DBLP:conf/icra/WeiSL021} &   4.48 & 3.28 & 3.15 & 3.97 & 4.51 & 4.21     \\
   \rowcolor{Gainsboro!60}\textsc{Greedy Box Seeker}  & \textbf{7.73} & \textbf{8.02} & \textbf{7.50} & \textbf{12.20} & \textbf{14.29} & \textbf{12.30}   \\
\bottomrule
\end{tabular}
}
\vspace{-4ex}
\end{table}

Table \ref{tab:6-class-all} presents a comparative analysis of methods on \textcolor{seal}{$\S$ \textsc{Setting} 1} including \textsc{Top-down Projection}, \textsc{Top-SelfTrain}, \textsc{Top-Clustering} and the proposed \textsc{Find N' Propagate}.  The term `\textsc{Base}' refers to the performance of backbone detectors pre-trained on 6 base classes. `CLIP$^*$' indicates the \textsc{Top-down} variant with the class-agnostic detector, which exhibits performance parallel to the base pretrained models. Table \ref{tab:3-class} reports the results under the challenging \textcolor{seal}{$\S$ \textsc{Setting} 2} with 7 novel classes. In this setting, the pretrained \textsc{Base} model has access to a restricted vocabulary, limiting its proposal generalization. 

\noindent \textbf{Compared with learning-free \textsc{Top-down}.}  Table \ref{tab:6-class-all} reveals that most \textsc{Top-down} approaches show an increase of up to 6\% in $\text{AP}_{\text{N}}$, albeit at the cost of a slight decrease in mAP compared to the pretrained base model. GLIP significantly outperforms CLIP and MaskCLIP in zero-shot detection, notably in detecting motorcycles. Models using logit fusion (indicated by -f) retain the base model's proficiency on known classes, with up to 19.6\% increase in mAP. CenterPoint tends to exhibit inferior regression quality, leading to a 1\% to 25\% drop in mAP. In contrast, our \textsc{Find n' Propagate} approach greatly outperforms all \textsc{Top-down} variants, elevating AP for novel classes by an impressive 110\% to 397\% using GLIP and OWL-ViT. For \textcolor{seal}{$\S$ \textsc{Setting} 2} results only our \textsc{Find n' Propagate} approach can achieve a reasonable AP$_{\text{N}}$ of 18\%. NDS is slightly lower due to a lack of velocity supervision in proposals mined from Box Seeker.

\noindent \textbf{Does Bottom-up OV-3D have higher recall for novels?} Table \ref{tab:6-class-all} reports the average recall (\textcolor{seal}{AR$_\text{N}$}) rate of novel classes in the last column. \textsc{Top-down} methods will have the same AR$_{\text{N}}$ for different VLMs since the proposals are the same up to label and score. Compared with \textsc{Top-down Self-train}, our \textsc{Find n' Propagate} approach gains an absolute increase of 20.7\% on AR$_{\text{N}}$ over \textsc{Top-down Self-train}, evidencing its effectiveness at discovering more novel instances. The proposed \textsc{Top-down Clustering} method cannot regress better proposals as objects can be vastly distributed and do not form densely packed and isolated clusters. Despite having access to the segmentation labels from CLIP2Scene, there are 6 classes with zero AP for HDBScan, and 7 for DBScan; both having a mAP less than 1\%.


\noindent \textbf{Compared with \textsc{Top-down Self-train}.} As shown in Tab. \ref{tab:6-class-all}, with the aid of self-training, \textsc{Top-down Self-train} with CLIP improves AP$_{\text{N}}$ from 3.1\% (learning-free) to 6.46\%. Unfortunately, for classes with distinct shapes like traffic cones, the improvement is negligible. Notably, \textsc{Find n' Propagate} consistently outperforms \textsc{Top-down Self-train}, enhancing AP$_{\text{N}}$ by 420\%. \textsc{Top-down Self-train} with GLIP is not able to drastically improve on its learning-free proposals, with AP$_{\text{N}}$ dropping by 16\%.


\noindent \textbf{Greedy Box Seeker v.s \textsc{Bottom-up Weakly-supervised}.} In Table \ref{tab:kitti_proposal}, we compare the proposed Greedy Box Seeker with the Baseline IV - Bottom-up Weakly-supervised approach FGR \cite{DBLP:conf/icra/WeiSL021}. The rectangular shape bias of FGR leads it to produce poor regressions for non-rectangular classes whilst our Greedy Box Seeker can regress better proposals for Cyclist and Pedestrian at all difficulties, having increased in AP$_{\text{3D}}$ by at least 73\%. 

\noindent \textbf{Greedy Box Seeker v.s. OpenSight \cite{DBLP:journals/corr/abs-2312-08876}.} We compare our proposal quality with a concurrent counterpart OpenSight \cite{DBLP:journals/corr/abs-2312-08876} on the nuScenes dataset, as shown in Table \ref{tab:0-class-all}. In \textcolor{seal}{$\S$ \textsc{Setting} 3}), we study the quality of boxes generated by OpenSight and by our Greedy Box Seeker without requiring any 3D annotations. We find that our proposals drastically improve the recovery of novel instances by 97\% AP$_{\text{N}}$, attesting to the flexibility of our box seeker.
\vspace{-2ex}
\subsection{Ablation Study and Visualisation}\label{sec:ablation}\vspace{-1ex} In this subsection, we study the impact of the proposed Greedy Box Oracle and two simulators. Due to the space limit, more ablation studies are provided in the supplementary material.

\noindent \textbf{Impact of Geometry Simulator.} We study the impact of the proposed simulators in Table \ref{tab:copy_and_paste_study} under \textcolor{seal}{$\S$ \textsc{Setting} 1}, exemplifying the need to sample extra novel instances of varying geometric characteristics during training. We experiment with different augmentation methods for novel instances, starting with GT Sampling, and then introduce modules from the Geometry Simulator. We find that with GT Sampling there is not enough geometric variance in the objects as it is pasting the already found objects without modification. Adding deviations to the pasted location increases AR$_{\text{N}}$ by 6\% absolute and produces best AP$_{\text{B}}$ by a margin of 1\%, however, decreases AP$_{\text{N}}$ by 8\%. Relative to translation alone, the proposed simulator with translation and rotation variance increases AP$_{\text{N}}$ by 73\%, benefiting from both orientation and spatial transformation. 

\noindent \textbf{Impact of Density Simulator.}  As reported in Table \ref{tab:copy_and_paste_study}, the proposed density simulator (row 2 and row 4) effectively emulates sparse versions of nearby objects with the same geometry. Compared with the standard GT sampling (row 1), the Transfusion detector trained with density simulator enabled gains 42\% of the detection precision of novel classes AP$_{\text{N}}$ and 15.6\% of recall AR$_{\text{N}}$ with the dropout rate $p_{\text{drop}} = 0.2$ under \textcolor{seal}{$\S$ \textsc{Setting} 1}). 

\noindent \textbf{Impact of Greedy Box Oracle.}
The importance of density ranker and multi-view alignment in filtering out noisy novel instances is underscored in our study. Table \ref{tab:criterion_study} presents a benchmark comparison of these two quality control criteria. View alignment, while effective, can inadvertently validate false proposals that are too far away due to the 2D box being too small, performing 6\% worse in AP$_{\text{N}}$. The combined use of both criteria yields optimal performance.

\begin{figure*}[t]
    \centering
    \includegraphics[width=1\linewidth]{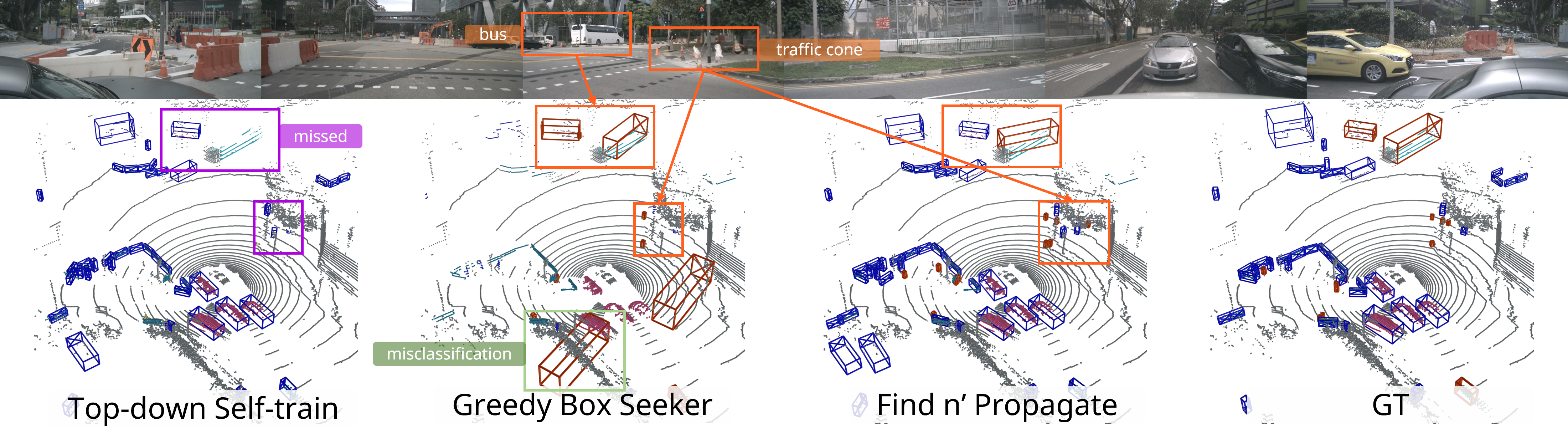}\vspace{-2ex}
    \caption{\textbf{Visualisation of open-vocabulary 3D detection results (\textcolor{seal}{$\S$ \textsc{Setting} 1}).}\vspace{-2ex} }\label{fig:vis-frustum}
\end{figure*}

\noindent\textbf{Visualisation.} Beyond the quantitative analysis, we further provide visualisations of 3D object detection results with the Transfusion backbone under  \textcolor{seal}{$\S$ \textsc{Setting} 1}), including the (1) \textsc{Top-down Self-train} model, (2) searched boxes from the proposed Greedy Box Seeker (2) the full version of the proposed \textsc{Find n' Propagate} approach and (4) ground-truth as shown in Fig. \ref{fig:vis-frustum}. It shows that our \textsc{Find n' Propagate} model can capture more unseen objects with varying sizes (\textit{e.g.}, traffic cones and buses highlighted in \textcolor{orange}{orange} boxes).

\begin{table}[t]
\centering
\begin{minipage}[t]{.71\linewidth}
    \caption{\textbf{Ablation study on the proposed geometry and density simulators (\textcolor{seal}{$\S$ \textsc{Setting} 1}).}\vspace{-2ex}}
    \label{tab:copy_and_paste_study}
    \centering
    \resizebox{1\linewidth}{!}{%
    \begin{tabular}{lcccccccc}
    \toprule
      Method & \multicolumn{1}{l}{$\sigma_{xyz}$} & \multicolumn{1}{l}{$\sigma_{\theta}$} & $p_{\text{drop}}$ &\multicolumn{1}{c}{mAP} & \multicolumn{1}{c}{NDS} & \multicolumn{1}{c}{AP$_\text{B}$} & \multicolumn{1}{c}{\textcolor{seal}{AP$_\text{N}$}} & \multicolumn{1}{c}{\textcolor{seal}{AR$_\text{N}$}} \\
    \midrule 
        GT Sampling \cite{DBLP:journals/sensors/YanML18} & -   & -  & - & 39.47 & 44.96 & 51.70 & 21.14 & 50.59 \\
        Ours & 0.0   & 0.0            & 0.2 & 43.85 & 47.03 & 53.07 & 30.00 & 52.61 \\
        Ours & 1.0   & 0.0            & 0.0 & 39.68 & 44.77 & \textbf{53.19} & 19.41 & 56.40 \\
        Ours & 1.0   &$\frac{\pi}{4}$ & 0.2 & \textbf{44.95} & \textbf{47.87} & 52.48 & \textbf{33.65} & \textbf{58.46} \\
    \bottomrule
\end{tabular}}\vspace{-4ex}
\end{minipage}
\hfill
\begin{minipage}[t]{.28\linewidth}
    \centering
    \caption{\textbf{Ablation study on the Greedy Box Oracle.}\vspace{-2ex}}
    \label{tab:criterion_study}
    \resizebox{1\linewidth}{!}{
    \begin{tabular}{cccccc}
    \toprule
      \multicolumn{1}{l}{$\mathfrak{C}_1$} & \multicolumn{1}{l}{$\mathfrak{C}_2$} & \multicolumn{1}{c}{NDS} & \multicolumn{1}{c}{\textcolor{seal}{AP$_\text{N}$}}& \multicolumn{1}{c}{\textcolor{seal}{AR$_\text{N}$}} \\
    \midrule 
       \checkmark &  & 20.69 & 15.97 & 24.08 \\
         & \checkmark  &  21.02 & 15.05 & 21.82 \\
       \checkmark & \checkmark  &  \textbf{22.40} & \textbf{16.72} & \textbf{26.70}  \\
       
    \bottomrule
    \end{tabular}}
\end{minipage}
\end{table}

\vspace{-2ex}
\section{Conclusion and Discussion}\vspace{-2ex} In our study, we extensively explored and benchmarked four primary baselines for open-vocabulary learning in 3D object detection. Our findings indicate that the proposed \textsc{Find n' Propagate} approach is more effective, particularly in detecting novel instances of varying sizes, thereby addressing the inherent biases of camera-based methods. However, limitations include suboptimal recall and difficulties in identifying objects across multiple views. Future work will focus on enhancing 2D fusion techniques and incorporating temporal constraints.
\vspace{-2ex}
\subsubsection*{Acknowledgements.} 
This research is partially supported by the Australian Research Council (DE240100105, DP240101814, DP230101196); JST Moonshot R\&D Grant Number JPMJPS2011, CREST Grant Number JPMJCR2015 and Basic Research Grant (Super AI) of Institute for AI and Beyond of the University of Tokyo. 


%
%
\bibliographystyle{splncs04}
\bibliography{main}

\begin{thebibliography}{10}
\providecommand{\url}[1]{\texttt{#1}}
\providecommand{\urlprefix}{URL }
\providecommand{\doi}[1]{https://doi.org/#1}

\bibitem{DBLP:conf/iros/AhmedTCMW18}
Ahmed, S.M., Tan, Y.Z., Chew, C., Mamun, A.A., Wong, F.S.: Edge and corner detection for unorganized 3d point clouds with application to robotic welding. In: International Conference on Intelligent Robots and Systems (IROS). pp. 7350--7355 (2018)

\bibitem{DBLP:conf/cvpr/BaiHZHCFT22}
Bai, X., Hu, Z., Zhu, X., Huang, Q., Chen, Y., Fu, H., Tai, C.: Transfusion: Robust lidar-camera fusion for 3d object detection with transformers. In: {IEEE/CVF} Conference on Computer Vision and Pattern Recognition (CVPR). pp. 1080--1089. {IEEE} (2022)

\bibitem{DBLP:conf/cvpr/CaesarBLVLXKPBB20}
Caesar, H., Bankiti, V., Lang, A.H., Vora, S., Liong, V.E., Xu, Q., Krishnan, A., Pan, Y., Baldan, G., Beijbom, O.: nuscenes: {A} multimodal dataset for autonomous driving. In: {IEEE/CVF} Conference on Computer Vision and Pattern Recognition (CVPR). pp. 11618--11628 (2020)

\bibitem{DBLP:conf/pakdd/CampelloMS13}
Campello, R.J.G.B., Moulavi, D., Sander, J.: Density-based clustering based on hierarchical density estimates. In: Pei, J., Tseng, V.S., Cao, L., Motoda, H., Xu, G. (eds.) Advances in Knowledge Discovery and Data Mining (PAKDD). vol.~7819, pp. 160--172. Springer (2013)

\bibitem{chen2023clip2scene}
Chen, R., Liu, Y., Kong, L., Zhu, X., Ma, Y., Li, Y., Hou, Y., Qiao, Y., Wang, W.: Clip2scene: Towards label-efficient 3d scene understanding by clip. In: IEEE/CVF Conference on Computer Vision and Pattern Recognition (CVPR). pp. 7020--7030 (2023)

\bibitem{DBLP:conf/iccv/ChenL0BH23}
Chen, Z., Luo, Y., Wang, Z., Baktashmotlagh, M., Huang, Z.: Revisiting domain-adaptive 3d object detection by reliable, diverse and class-balanced pseudo-labeling. In: International Conference on Computer Vision (ICCV). pp. 3691--3703 (2023)

\bibitem{DBLP:conf/nips/DengQNFZA21}
Deng, B., Qi, C.R., Najibi, M., Funkhouser, T.A., Zhou, Y., Anguelov, D.: Revisiting 3d object detection from an egocentric perspective. In: Advances in Neural Information Processing Systems (NeurIPS). pp. 26066--26079 (2021)

\bibitem{DBLP:conf/kdd/EsterKSX96}
Ester, M., Kriegel, H., Sander, J., Xu, X.: A density-based algorithm for discovering clusters in large spatial databases with noise. In: Simoudis, E., Han, J., Fayyad, U.M. (eds.) Conference on Knowledge Discovery and Data Mining (KDD). pp. 226--231. {AAAI} Press (1996)

\bibitem{DBLP:conf/cvpr/GeigerLU12}
Geiger, A., Lenz, P., Urtasun, R.: Are we ready for autonomous driving? the {KITTI} vision benchmark suite. In: {IEEE} Conference on Computer Vision and Pattern Recognition (CVPR). pp. 3354--3361. {IEEE} Computer Society (2012)

\bibitem{gu2021open}
Gu, X., Lin, T.Y., Kuo, W., Cui, Y.: Open-vocabulary detection via vision and language knowledge distillation. arXiv preprint arXiv:2104.13921  (2021)

\bibitem{DBLP:conf/corl/HoustonZBYCJOIO20}
Houston, J., Zuidhof, G., Bergamini, L., Ye, Y., Chen, L., Jain, A., Omari, S., Iglovikov, V., Ondruska, P.: One thousand and one hours: Self-driving motion prediction dataset. In: Kober, J., Ramos, F., Tomlin, C.J. (eds.) Conference on Robot Learning, (CoRL). Proceedings of Machine Learning Research, vol.~155, pp. 409--418. {PMLR} (2020)

\bibitem{DBLP:journals/corr/abs-2312-07530}
Huang, K., Tsai, Y., Yang, M.: Weakly supervised 3d object detection via multi-level visual guidance. CoRR  \textbf{abs/2312.07530} (2023)

\bibitem{huang2023clip2point}
Huang, T., Dong, B., Yang, Y., Huang, X., Lau, R.W., Ouyang, W., Zuo, W.: Clip2point: Transfer clip to point cloud classification with image-depth pre-training. In: IEEE/CVF Conference on Computer Vision and Pattern Recognition (CVPR). pp. 22157--22167 (2023)

\bibitem{DBLP:journals/corr/abs-2309-00775}
Kim, D., Angelova, A., Kuo, W.: Contrastive feature masking open-vocabulary vision transformer. CoRR  \textbf{abs/2309.00775} (2023)

\bibitem{DBLP:conf/iclr/KuoCGPA23}
Kuo, W., Cui, Y., Gu, X., Piergiovanni, A.J., Angelova, A.: Open-vocabulary object detection upon frozen vision and language models. In: International Conference on Learning Representations (ICLR) (2023)

\bibitem{DBLP:conf/cvpr/LiZZYLZWYZHCG22}
Li, L.H., Zhang, P., Zhang, H., Yang, J., Li, C., Zhong, Y., Wang, L., Yuan, L., Zhang, L., Hwang, J., Chang, K., Gao, J.: Grounded language-image pre-training. In: {IEEE/CVF} Conference on Computer Vision and Pattern Recognition, {CVPR} 2022, New Orleans, LA, USA, June 18-24, 2022. pp. 10955--10965 (2022)

\bibitem{DBLP:conf/eccv/LiuQHQLTW22}
Liu, C., Qian, X., Huang, B., Qi, X., Lam, E.Y., Tan, S., Wong, N.: Multimodal transformer for automatic 3d annotation and object detection. In: Avidan, S., Brostow, G.J., Ciss{\'{e}}, M., Farinella, G.M., Hassner, T. (eds.) European Conference on Computer Vision (ECCV). vol. 13698, pp. 657--673. Springer (2022)

\bibitem{liu2023partslip}
Liu, M., Zhu, Y., Cai, H., Han, S., Ling, Z., Porikli, F., Su, H.: Partslip: Low-shot part segmentation for 3d point clouds via pretrained image-language models. In: IEEE/CVF Conference on Computer Vision and Pattern Recognition (CVPR). pp. 21736--21746 (2023)

\bibitem{liu2023segment}
Liu, Y., Kong, L., Cen, J., Chen, R., Zhang, W., Pan, L., Chen, K., Liu, Z.: Segment any point cloud sequences by distilling vision foundation models. In: Advances in Neural Information Processing Systems (NeurIPS) (2023)

\bibitem{DBLP:conf/cvpr/LuXWXTKZ23}
Lu, Y., Xu, C., Wei, X., Xie, X., Tomizuka, M., Keutzer, K., Zhang, S.: Open-vocabulary point-cloud object detection without 3d annotation. In: {IEEE/CVF} Conference on Computer Vision and Pattern Recognition (CVPR). pp. 1190--1199 (2023)

\bibitem{DBLP:conf/iccv/LuoCF0BH23}
Luo, Y., Chen, Z., Fang, Z., Zhang, Z., Baktashmotlagh, M., Huang, Z.: Kecor: Kernel coding rate maximization for active 3d object detection. In: International Conference on Computer Vision (ICCV). pp. 18233--18244 (2023)

\bibitem{DBLP:conf/iclr/LuoCWYHB23}
Luo, Y., Chen, Z., Wang, Z., Yu, X., Huang, Z., Baktashmotlagh, M.: Exploring active 3d object detection from a generalization perspective. In: International conference on machine learning (ICLR) (2023)

\bibitem{DBLP:journals/corr/abs-2310-16667}
Ma, C., Jiang, Y., Wen, X., Yuan, Z., Qi, X.: Codet: Co-occurrence guided region-word alignment for open-vocabulary object detection. CoRR  \textbf{abs/2310.16667} (2023)

\bibitem{DBLP:conf/nips/MaoNJLCLLY0LYXX21}
Mao, J., Niu, M., Jiang, C., Liang, H., Chen, J., Liang, X., Li, Y., Ye, C., Zhang, W., Li, Z., Yu, J., Xu, C., Xu, H.: One million scenes for autonomous driving: {ONCE} dataset. In: Vanschoren, J., Yeung, S. (eds.) Advances in Neural Information Processing Systems (NeurIPS) (2021)

\bibitem{DBLP:journals/ijcv/MaoSWL23}
Mao, J., Shi, S., Wang, X., Li, H.: 3d object detection for autonomous driving: {A} comprehensive survey. Int. J. Comput. Vis.  \textbf{131}(8),  1909--1963 (2023)

\bibitem{DBLP:conf/eccv/MengWZSGD20}
Meng, Q., Wang, W., Zhou, T., Shen, J., Gool, L.V., Dai, D.: Weakly supervised 3d object detection from lidar point cloud. In: Vedaldi, A., Bischof, H., Brox, T., Frahm, J. (eds.) European Conference on Computer Vision (ECCV). vol. 12358, pp. 515--531. Springer (2020)

\bibitem{DBLP:journals/corr/abs-2306-09683}
Minderer, M., Gritsenko, A.A., Houlsby, N.: Scaling open-vocabulary object detection. CoRR  \textbf{abs/2306.09683} (2023)

\bibitem{DBLP:journals/corr/abs-2205-06230}
Minderer, M., Gritsenko, A.A., Stone, A., Neumann, M., Weissenborn, D., Dosovitskiy, A., Mahendran, A., Arnab, A., Dehghani, M., Shen, Z., Wang, X., Zhai, X., Kipf, T., Houlsby, N.: Simple open-vocabulary object detection with vision transformers. CoRR  \textbf{abs/2205.06230} (2022)

\bibitem{DBLP:conf/iros/MontesLCD20}
Montes, H.A., Louedec, J.L., Cielniak, G., Duckett, T.: Real-time detection of broccoli crops in 3d point clouds for autonomous robotic harvesting. In: International Conference on Intelligent Robots and Systems (IROS). pp. 10483--10488 (2020)

\bibitem{Peng2023OpenScene}
Peng, S., Genova, K., Jiang, C.M., Tagliasacchi, A., Pollefeys, M., Funkhouser, T.: Openscene: 3d scene understanding with open vocabularies. In: IEEE/CVF Conference on Computer Vision and Pattern Recognition (CVPR) (CVPR) (2023)

\bibitem{DBLP:conf/cvpr/QiLWSG18}
Qi, C.R., Liu, W., Wu, C., Su, H., Guibas, L.J.: Frustum pointnets for 3d object detection from {RGB-D} data. In: {IEEE} Conference on Computer Vision and Pattern Recognition (CVPR). pp. 918--927. Computer Vision Foundation / {IEEE} Computer Society (2018)

\bibitem{DBLP:conf/icml/RadfordKHRGASAM21}
Radford, A., Kim, J.W., Hallacy, C., Ramesh, A., Goh, G., Agarwal, S., Sastry, G., Askell, A., Mishkin, P., Clark, J., Krueger, G., Sutskever, I.: Learning transferable visual models from natural language supervision. In: International Conference on Machine Learning (ICML). vol.~139, pp. 8748--8763. {PMLR} (2021)

\bibitem{radford2021learning}
Radford, A., Kim, J.W., Hallacy, C., Ramesh, A., Goh, G., Agarwal, S., Sastry, G., Askell, A., Mishkin, P., Clark, J., et~al.: Learning transferable visual models from natural language supervision. In: International conference on machine learning (ICLR). pp. 8748--8763. PMLR (2021)

\bibitem{NIPS2015_14bfa6bb}
Ren, S., He, K., Girshick, R., Sun, J.: Faster r-cnn: Towards real-time object detection with region proposal networks. In: Advances in Neural Information Processing Systems (NeurIPS). vol.~28. Curran Associates, Inc. (2015)

\bibitem{DBLP:journals/corr/abs-2401-06542}
Song, Z., Liu, L., Jia, F., Luo, Y., Zhang, G., Yang, L., Wang, L., Jia, C.: Robustness-aware 3d object detection in autonomous driving: {A} review and outlook. CoRR  \textbf{abs/2401.06542} (2024)

\bibitem{DBLP:conf/cvpr/SunKDCPTGZCCVHN20}
Sun, P., Kretzschmar, H., Dotiwalla, X., Chouard, A., Patnaik, V., Tsui, P., Guo, J., Zhou, Y., Chai, Y., Caine, B., Vasudevan, V., Han, W., Ngiam, J., Zhao, H., Timofeev, A., Ettinger, S., Krivokon, M., Gao, A., Joshi, A., Zhang, Y., Shlens, J., Chen, Z., Anguelov, D.: Scalability in perception for autonomous driving: Waymo open dataset. In: IEEE Conference on Computer Vision and Pattern Recognition (CVPR). pp. 2443--2451 (2020)

\bibitem{DBLP:conf/cvpr/TaoHQ0S23}
Tao, R., Han, W., Qiu, Z., Xu, C., Shen, J.: Weakly supervised monocular 3d object detection using multi-view projection and direction consistency. In: {IEEE/CVF} Conference on Computer Vision and Pattern Recognition (CVPR). pp. 17482--17492. {IEEE} (2023)

\bibitem{DBLP:conf/eccv/WangLGD20}
Wang, J., Lan, S., Gao, M., Davis, L.S.: Infofocus: 3d object detection for autonomous driving with dynamic information modeling. In: European Conference on Computer Vision (ECCV). vol. 12355, pp. 405--420 (2020)

\bibitem{DBLP:journals/sensors/WangLSLZSQT19}
Wang, L., Li, R., Sun, J., Liu, X., Zhao, L., Seah, H.S., Quah, C.K., Tandianus, B.: Multi-view fusion-based 3d object detection for robot indoor scene perception. Sensors  \textbf{19}(19), ~4092 (2019)

\bibitem{DBLP:conf/icra/WeiSL021}
Wei, Y., Su, S., Lu, J., Zhou, J.: {FGR:} frustum-aware geometric reasoning for weakly supervised 3d vehicle detection. In: {IEEE} International Conference on Robotics and Automation (ICRA). pp. 4348--4354. {IEEE} (2021)

\bibitem{wu2023cora}
Wu, X., Zhu, F., Zhao, R., Li, H.: Cora: Adapting clip for open-vocabulary detection with region prompting and anchor pre-matching. In: IEEE/CVF Conference on Computer Vision and Pattern Recognition (CVPR) (CVPR). pp. 7031--7040 (2023)

\bibitem{xue2023ulip}
Xue, L., Gao, M., Xing, C., Mart{\'\i}n-Mart{\'\i}n, R., Wu, J., Xiong, C., Xu, R., Niebles, J.C., Savarese, S.: Ulip: Learning a unified representation of language, images, and point clouds for 3d understanding. In: IEEE/CVF Conference on Computer Vision and Pattern Recognition (CVPR). pp. 1179--1189 (2023)

\bibitem{DBLP:journals/sensors/YanML18}
Yan, Y., Mao, Y., Li, B.: {SECOND:} sparsely embedded convolutional detection. Sensors  \textbf{18}(10), ~3337 (2018). \doi{10.3390/S18103337}, \url{https://doi.org/10.3390/s18103337}

\bibitem{DBLP:conf/cvpr/YaoHLXZLX23}
Yao, L., Han, J., Liang, X., Xu, D., Zhang, W., Li, Z., Xu, H.: Detclipv2: Scalable open-vocabulary object detection pre-training via word-region alignment. In: {IEEE/CVF} Conference on Computer Vision and Pattern Recognition (CVPR). pp. 23497--23506. {IEEE} (2023)

\bibitem{DBLP:conf/nips/YaoHWLX0LXX22}
Yao, L., Han, J., Wen, Y., Liang, X., Xu, D., Zhang, W., Li, Z., Xu, C., Xu, H.: Detclip: Dictionary-enriched visual-concept paralleled pre-training for open-world detection. In: Advances in Neural Information Processing Systems (NeurIPS) (2022)

\bibitem{DBLP:conf/cvpr/YinZK21}
Yin, T., Zhou, X., Kr{\"{a}}henb{\"{u}}hl, P.: Center-based 3d object detection and tracking. In: {IEEE/CVF} Conference on Computer Vision and Pattern Recognition (CVPR). pp. 11784--11793 (2021)

\bibitem{zang2022open}
Zang, Y., Li, W., Zhou, K., Huang, C., Loy, C.C.: Open-vocabulary detr with conditional matching (2022)

\bibitem{zareian2021open}
Zareian, A., Rosa, K.D., Hu, D.H., Chang, S.F.: Open-vocabulary object detection using captions. In: IEEE/CVF Conference on Computer Vision and Pattern Recognition (CVPR) (CVPR). pp. 14393--14402 (2021)

\bibitem{zeng2023clip2}
Zeng, Y., Jiang, C., Mao, J., Han, J., Ye, C., Huang, Q., Yeung, D.Y., Yang, Z., Liang, X., Xu, H.: Clip2: Contrastive language-image-point pretraining from real-world point cloud data. In: IEEE/CVF Conference on Computer Vision and Pattern Recognition (CVPR). pp. 15244--15253 (2023)

\bibitem{DBLP:journals/corr/abs-2303-15343}
Zhai, X., Mustafa, B., Kolesnikov, A., Beyer, L.: Sigmoid loss for language image pre-training. CoRR  \textbf{abs/2303.15343} (2023)

\bibitem{DBLP:conf/nips/ZhangZ00LDWYHG22}
Zhang, H., Zhang, P., Hu, X., Chen, Y., Li, L.H., Dai, X., Wang, L., Yuan, L., Hwang, J., Gao, J.: Glipv2: Unifying localization and vision-language understanding. In: Advances in Neural Information Processing Systems (NeurIPS) (2022)

\bibitem{DBLP:journals/corr/abs-2312-08876}
Zhang, H., Xu, J., Tang, T., Sun, H., Yu, X., Huang, Z., Yu, K.: Opensight: {A} simple open-vocabulary framework for lidar-based object detection. CoRR  \textbf{abs/2312.08876} (2023)

\bibitem{Zhang_2022_CVPR}
Zhang, R., Guo, Z., Zhang, W., Li, K., Miao, X., Cui, B., Qiao, Y., Gao, P., Li, H.: Pointclip: Point cloud understanding by clip. In: IEEE/CVF Conference on Computer Vision and Pattern Recognition (CVPR). pp. 8552--8562 (June 2022)

\bibitem{DBLP:conf/eccv/ZhaoZSZKSCM22}
Zhao, S., Zhang, Z., Schulter, S., Zhao, L., Kumar, B.G.V., Stathopoulos, A., Chandraker, M., Metaxas, D.N.: Exploiting unlabeled data with vision and language models for object detection. In: European Conference on Computer Vision (ECCV). vol. 13669, pp. 159--175. Springer (2022)

\bibitem{DBLP:conf/cvpr/ZhongYZLCLZDYLG22}
Zhong, Y., Yang, J., Zhang, P., Li, C., Codella, N., Li, L.H., Zhou, L., Dai, X., Yuan, L., Li, Y., Gao, J.: Regionclip: Region-based language-image pretraining. In: {IEEE/CVF} Conference on Computer Vision and Pattern Recognition (CVPR). pp. 16772--16782. {IEEE} (2022)

\bibitem{DBLP:conf/eccv/ZhouLD22}
Zhou, C., Loy, C.C., Dai, B.: Extract free dense labels from {CLIP}. In: European Conference on Computer Vision (ECCV). vol. 13688, pp. 696--712. Springer (2022)

\bibitem{DBLP:conf/eccv/ZhouGJKM22}
Zhou, X., Girdhar, R., Joulin, A., Kr{\"{a}}henb{\"{u}}hl, P., Misra, I.: Detecting twenty-thousand classes using image-level supervision. In: European Conference on Computer Vision (ECCV). vol. 13669, pp. 350--368. Springer (2022)

\bibitem{zhu2023pointclip}
Zhu, X., Zhang, R., He, B., Guo, Z., Zeng, Z., Qin, Z., Zhang, S., Gao, P.: Pointclip v2: Prompting clip and gpt for powerful 3d open-world learning. In: IEEE/CVF Conference on Computer Vision and Pattern Recognition (CVPR). pp. 2639--2650 (2023)

\end{thebibliography}
\end{document}


\title{Supplementary Material for Find n' Propagate: Open-Vocabulary 3D Object Detection in Urban Environments}
\titlerunning{Find n' Propagate Supplementary Material}

\author{Djamahl Etchegaray\inst{1}\orcidlink{0009-0000-1154-5389} \and
Zi Huang\inst{1}\orcidlink{0000-0002-9738-4949} \and
Tatsuya Harada\inst{2}\orcidlink{0000-0002-3712-3691} \and Yadan Luo \thanks{Correspondence to Yadan Luo (y.luo@uq.edu.au)}\inst{1}\orcidlink{0000-0001-6272-2971}}

\authorrunning{D.~Etchegaray et al.}

\institute{UQMM Lab, University of Queensland, Brisbane, Australia
\\
\and
The University of Tokyo, Tokyo, Japan. \\ \email{\{uqdetche, helen.huang, y.luo\}@uq.edu.au, harada@mi.t.u-tokyo.ac.jp}}

\maketitle

\section*{Overview}
This supplementary material provides complementary details to better understand the main paper. A brief overview of the sections is available below:

\begin{itemize}
    \item \ref{sec:visualisation}: Visualisation of Detection Results under \textcolor{seal}{\S\textsc{Setting 1}} and \textcolor{seal}{\S\textsc{Setting 3}}
    \item \ref{sec:implementation_details}: Further Implementation Details
    \begin{itemize}
        \item \ref{sec:st_filtering}: Filtering Strategies in Iterative Self-Training
    \end{itemize}
    \item \ref{sec:more_ablations}: More Ablation Studies
    \begin{itemize}
        \item \ref{sec:frustum_proposals}: \textsc{Greedy Box Oracle}
        \item \ref{sec:self_training}: Self-Training Objectives
        \item \ref{sec:topdown}: \textsc{Top-down Projection}
        \item \ref{sec:clip2scene}: Discussions on CLIP2Scene Zero-shot Results
    \end{itemize}
    
\end{itemize}

\section{Visualisation}\label{sec:visualisation}

\begin{figure*}[tp]
    \centering
    \begin{subfigure}{\textwidth}
      \includegraphics[width=\linewidth]{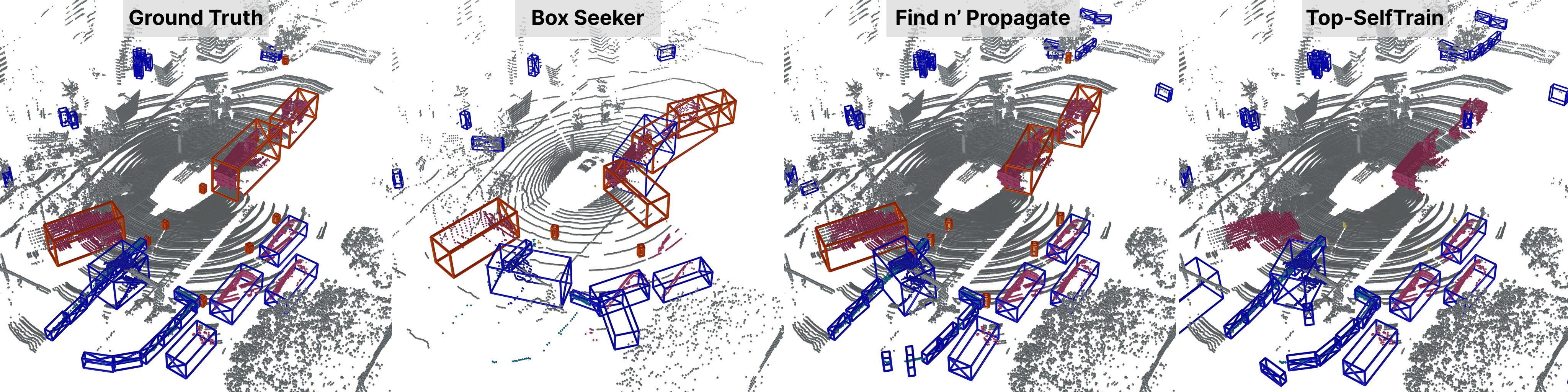}
      \caption{None of the novel instances can be found in this scene with \textsc{Top-Down self-train}, but \textsc{Box Seeker} and \textsc{Find n' Propagate} learning can sufficiently localise them. The frustum proposals can localise many novel and base objects, however, the localisation is noisy as can be seen with the varied orientations (camera facing bias). The \textsc{Find n' Propagate} model is able to improve the orientation \& position noise from the frustum proposals.}
    \end{subfigure}
    \begin{subfigure}{\textwidth}
      \includegraphics[width=\linewidth]{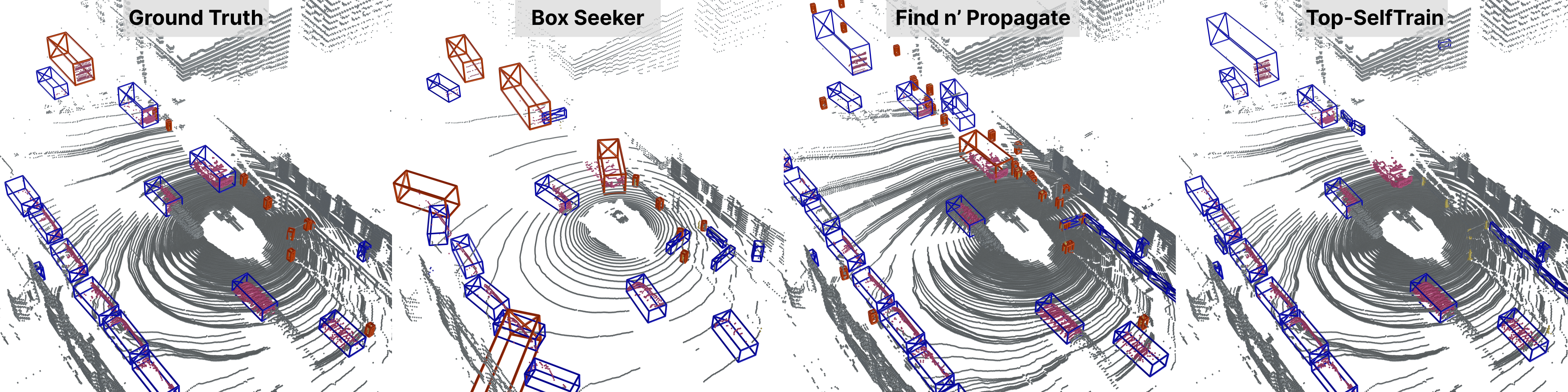}
      \caption{In this scene the recall is also significantly better for \textsc{Find n' Propagate}, however, the detector has produced a lot of traffic cones to maximise their recall. We can see that a number of them are true positives. Unfortunately, the frustum proposals contain a lot of misclassification errors from GLIP, where many base classes are incorrectly mapped to novel classes. }
    \end{subfigure}
    \begin{subfigure}{\textwidth}
      \includegraphics[width=\linewidth]{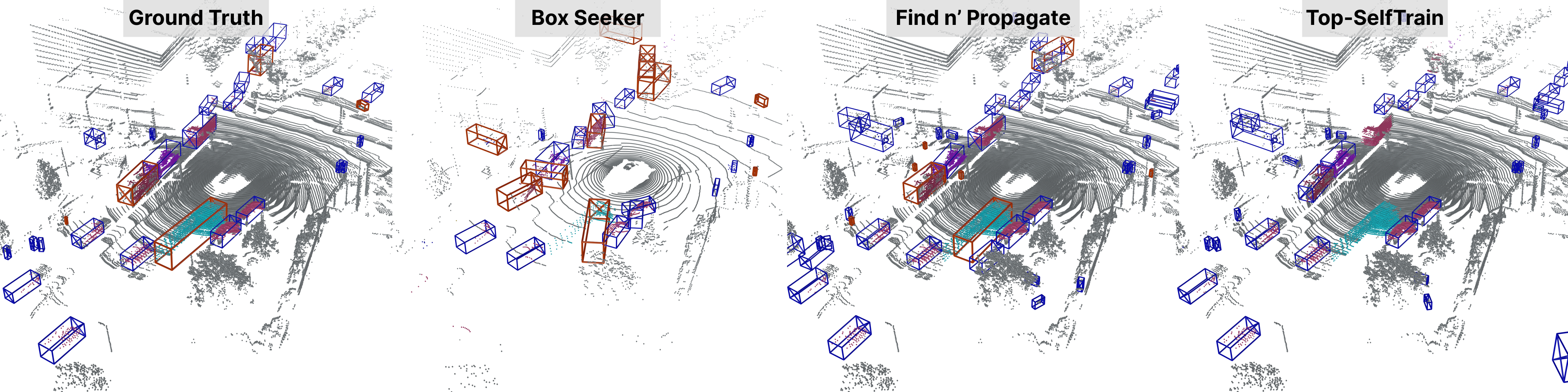}
      \caption{\textsc{Top-down self-train} can suffer from poor base performance, here \textsc{Top-selftrain} with CLIP is unable to find the large vehicle towards the centre of the point cloud. Like previous visualisations, no novels are proposed for this scene by \textsc{Top-down Self-train}. }
    \end{subfigure}
    \caption{\textbf{Visualisation of the Ground Truth, \textsc{Greedy Box Seeker}, \textsc{Find n' Propagate} and \textsc{Top-down Self-Train}.} Predicted boxes are coloured \textcolor{blue}{blue} for base classes and \textcolor{red}{red} for novel classes, under \textcolor{seal}{$\S$ \textsc{Setting} 1}. Points are coloured based on the class of the ground truth box they are in to help highlight which objects have been missed, if they do not fall in any ground-truth box their colour is grey. The \textcolor{blue}{base} ground truth and \textcolor{red}{novel} \textsc{Box Seeker} proposals are utilised for \textsc{Find n' Propagate} self-training. The \textcolor{blue}{base} \textsc{Seeker} proposals are shown for visualisation only, as in reality \textsc{Box Seeker} is under \textcolor{seal}{$\S$ \textsc{Setting} 3} and has no base classes it has learnt from. The effectiveness of our \textsc{Find n' Propagate} instance learning is highlighted, as \textsc{Top-selftrain} misses many novel and some base objects but ours is able to improve the recall drastically without poor regression quality. } 
    \label{fig:6cls-vis}
\end{figure*}

To achieve a better understanding of the proposed approach, we provide more visualisations of qualitative analysis on the derived \textsc{Box Seeker} (column 2), \textsc{Find n' Propagate} (column 3) and \textsc{Top-down self-training} on \textcolor{seal}{\S\textsc{Setting 1}} in Fig. \ref{fig:6cls-vis}. The top row gives the corresponding multi-view images of the studied point cloud. The base classes are highlighted in blue while the novel ones are colored in red. Points are coloured based on the class of the ground truth box they are in to help highlight which objects have been missed, if they do not fall in any ground-truth box their colour is grey. The figure shows the significant increase in the novel recall by using \textsc{Find n' Propagate} over a strong baseline \textsc{Top-down self-train}. Further, it establishes the number of new objects that were able to be found/refined in self-training. The detailed comparisons can be found in the captions of each figure.

\noindent \textbf{\textcolor{seal}{\S\textsc{Setting 3}}.} We previously evaluated our novel proposal generation on nuScenes, demonstrating the effectiveness on both base and novel classes. Visualisations of these proposals for all classes are available in Fig. \ref{fig:6cls-vis} (column 2). Further to our self-training experiments with \textcolor{seal}{\S\textsc{Setting 1}} and \textcolor{seal}{\S\textsc{Setting 2}}, we experimented with training on \textsc{Setting} 3 with the \textsc{Box Seeker}. Regrettably, our findings indicate that the proposed self-training technique struggled to effectively learn from the noisy proposals, primarily attributing this challenge to the inherent difficulty in assimilating semantic information from the noisy center heatmap. While some common objects were successfully identified, the majority remained elusive, resulting in a notably low precision. We intend to tackle this issue in our future research endeavors.

\section{Implementation Details}\label{sec:implementation_details}

In our study, we implement both \textsc{Top-up} and \textsc{Bottom-down} OV-3D approaches in the OpenPCDet\footnote{\url{https://github.com/open-mmlab/OpenPCDet}} codebase. The source code is available in the supplementary material for reference. We use Transfusion \cite{DBLP:conf/cvpr/BaiHZHCFT22} and Centerpoint \cite{DBLP:conf/cvpr/YinZK21} as the 3D detection backbones with the default hyperparameters. For the implementation of logit fusion in the \textsc{Top-down} approaches, the threshold 
 $\gamma$ is set to 0.2. For the search space in \textsc{Greedy Box Seeker}, we configure the number of intervals for orientation  $k_{o}$ at $10$, scale $k_s$ at 4, and depth $k_d$ also at 4. In augmentation, the memory bank size $|\mathcal{Q}|$ for each class is 60. The training batch sizes are consistently fixed to 8. The Adam optimizer is adopted with a learning rate initiated as 0.001, and scheduled by the OneCycle scheduler. Different from \cite{DBLP:conf/cvpr/BaiHZHCFT22,DBLP:conf/cvpr/YinZK21}, we do not disable GT sampling for the last 5 epochs of training, as we found it would deteriorate the novel performance.

 
 \noindent\textbf{Top-down Clustering.} For HDBScan we use a minimal cluster size of 15, and for DBScan an epsilon of 1.5. We utilise the sklearn package \cite{scikit-learn} in Python for extracting clusters followed by box estimation from \cite{DBLP:conf/cvpr/LuXWXTKZ23}.



\subsection{Filtering Strategies in Iterative Self-training}\label{sec:st_filtering}
In this section, we elaborate on the filtering strategies in the last self-training step of the proposed \textsc{Find n' Propagate} framework. During iterative self-training it is necessary to filter and combine objects from multiple sources. Sources include \textsc{Box Seeker} proposals, self-training pseudo boxes, and \textsc{Remote Propagator} boxes to be pasted. Firstly, for all the sources we must remove any that overlap with GT, as detailed below in Filtering Novels. Secondly, once we filter out any novel boxes that overlap with GT, we must make sure that none overlap with each other. We utilise standard non-max suppression (NMS) with the pseudo scores to remove any boxes that overlap and keep only the ones that were well captured in self-training. Thirdly, we must remove novel proposals that are too close or too sparse, as these are likely to be low quality or false positives.    

\noindent \textbf{Filtering Novels.}  Specifically, we remove erroneous novel instances, $\{ \mathcal{Q}_i \}_{i=1}^{|\mathcal{Q}|}$, that overlap with GT base annotations, $\{ \mathfrak{B}_j^{\text{B}} \}_{j=1}^M$, by calculating the maximum IoU with any base GT, and removing those above a threshold, $\beta_{\text{overlap}}$:
\begin{equation*}
    \texttt{remove}(\mathcal{Q}_i) = \left( \texttt{max}_{k \in |\mathcal{Q}|}\texttt{IoU}(\mathfrak{B}_j^{\text{Base}}, \mathcal{Q}_i) \right) > \beta_{\text{overlap}}.
\end{equation*} We empirically set $\beta_{\text{overlap}} = 0.1$ to allow some overlap as the frustum proposals often have erroneous orientations. 


\noindent \textbf{Filtering Too Sparse and Too Close Objects.} During self-training, we maintain a queue of $|\mathcal{Q}|$ novel samples, which can be derived from both \textsc{Box Seeker} proposals and pseudo-labels. To update the queue for each batch, we rank boxes by their confidence, then filter out any that have \textit{too few} points or occur \textit{too close} to the ego vehicle. As we have already filtered novel instances that overlap with known GT annotations, then the high-confidence predictions should be unknown objects that are newly discovered.

\noindent \textbf{Combining \textsc{Seeker} Proposals and Self-Training Pseudos.} Iterative self-training must integrate proposals from \textsc{Box Seeker} and high-confidence pseudo-boxes from self-training. To do this, we concatenate them together and perform NMS. The \textsc{Box Seeker} utilises the VLM score as the confidence score since the quality control score is local to each frustum. As the VLM score is usually quite high, the self-training pseudo has to be very confident to replace the frustum proposal.




\section{More Ablation Studies}\label{sec:more_ablations}

Due to space limitations in the main paper, we include further ablations for all the proposed modules. We evaluate the effectiveness of the \textsc{Region VLM} (Sc. \ref{sec:region_vlm}), \textsc{Greedy Box Oracle} (Sec.  \ref{sec:frustum_proposals}): the values of the alignment criteria coefficient, search space depth quantiles and search space size sensitivity. Then, we study the design choices in the Sec. \ref{sec:self_training} self-training module, including loss normalisation/reweighting. Finally, we supply the details for our \textsc{Top-down} comparisons (Sec. \ref{sec:topdown}), providing the steps for fusing the labels from 3D proposals and VLM predictions. 

\subsection{\textsc{Region VLM}}\label{sec:region_vlm} The proposed approach is agnostic to region VLM choices. We compared the quality of the \textsc{Greedy Box Seeker} with various 2D prediction sources in Table \ref{tab:0-class-all}. The GT$_{\text{2D}}$ serves the upper bound of the performance. Cameras capture close and dense objects, missing many faraway ones whilst the GT$_{\text{2D}}$ captures every labelled object. However, 3D annotations are projected to generate GT$_{\text{2D}}$ and often do not tightly bound the 2D view of the object. Due to this, both OWL-ViT and GLIP can achieve \textit{better} results: both surpass GT for Cars and Buses classes and GLIP is better for Construction Vehicles, Trucks and Motorcycles. GLIP and OWL-ViT can achieve surprising performance, with GLIP being 28\% off the GT$_{\text{2D}}$ mAP, finding most nearby objects.

\begin{table*}[t]\caption{\textbf{Evaluation on nuScenes with 10 novel classes (\textcolor{seal}{$\S$ \textsc{Setting} 3}).}\vspace{-2ex}}
    \centering
    \resizebox{1\linewidth}{!}{
    \begin{tabular}{lc|  S[table-format=2.2]S[table-format=2.2]S[table-format=2.2]S[table-format=2.2]S[table-format=2.2]S[table-format=2.2] S[table-format=2.2]S[table-format=2.2]S[table-format=2.2]S[table-format=2.2]| S[table-format=2.2]S[table-format=2.2]}
    \toprule
      & &  \multicolumn{10}{c}{\textcolor{seal}{\textsc{Novel}}}     & \multicolumn{2}{c}{\textsc{Overall}}     \\
Box$_{\text{3D}}$ & VLM   &  \multicolumn{1}{l}{\textcolor{seal}{Car}} &\multicolumn{1}{l}{\textcolor{seal}{Const.}} &\multicolumn{1}{l}{\textcolor{seal}{Trai.}} & \multicolumn{1}{l}{\textcolor{seal}{Barr.}} & \multicolumn{1}{l}{\textcolor{seal}{Bic.}} & \multicolumn{1}{l}{\textcolor{seal}{Ped.}} &\multicolumn{1}{l}{\textcolor{seal}{Truck}} & \multicolumn{1}{l}{\textcolor{seal}{Bus}} & \multicolumn{1}{l}{\textcolor{seal}{Motor.}} & \multicolumn{1}{l}{\textcolor{seal}{Cone.}} & \multicolumn{1}{l}{\textcolor{seal}{AP$_{\text{N}}$}} & \multicolumn{1}{l}{NDS}   \\
\midrule %
     \rowcolor{Gainsboro!60}   & OWL  & \textbf{25.14} & 0.76 & 0.00 & 0.30 & 19.21 & 12.24 & 6.07 & 5.89 & 18.01 & 18.89 & 10.65 & 18.30 \\ 
    
   \rowcolor{Gainsboro!60}\multirow{-1}{*}{\textsc{Seeker}}   & GLIP  &  24.28 & \textbf{4.14} & \textbf{0.15} & \textbf{4.39} & \textbf{34.65} & \textbf{22.80} & \textbf{8.58} & \textbf{11.09} & \textbf{35.82} & \textbf{21.26} & \textbf{16.72} & \textbf{22.40} \\ 
    \rowcolor{Gainsboro!60}  & \textbf{GT}$_{\text{2D}}$  &  16.09 & 3.26 & 0.37 & 34.28 & 38.22 & 39.03 & 7.41 & 5.54 & 30.33 & 56.45 & 23.10 & 22.83 \\
 \bottomrule
    \end{tabular}
}
    \vspace{-4ex}
    \label{tab:0-class-all}
\end{table*}

\subsection{\textsc{Greedy Box Oracle}}\label{sec:frustum_proposals}
The following ablations discuss the effect of different frustum proposal generation components. For all settings, AP$_\text{B}$ is defined for 6 classes known where the evaluation is done directly on frustum proposals.

\begin{figure}[H]
    \centering
    \includegraphics[width=0.7\linewidth]{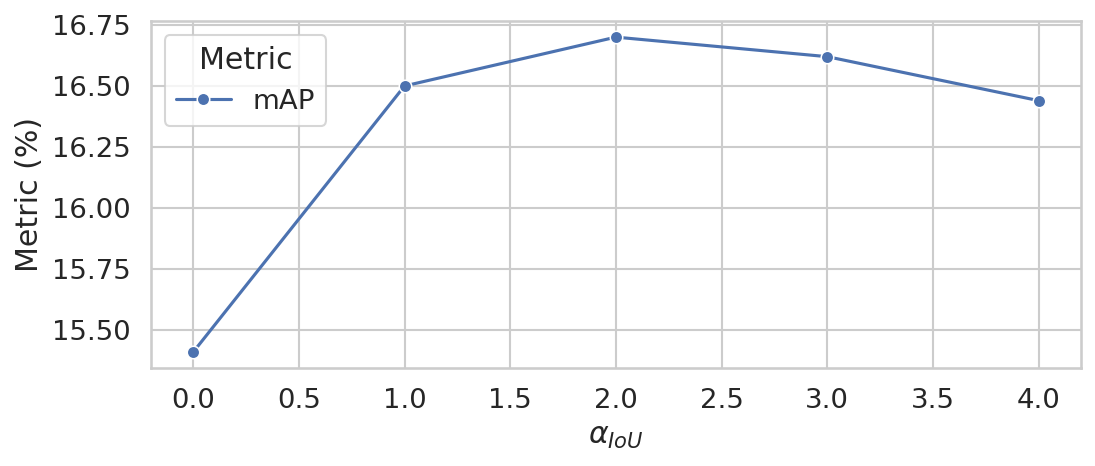}
    \caption{\textbf{Impact of Coefficient $\alpha_{\operatorname{IoU}}$} under \textcolor{seal}{$\S$ \textsc{Setting} 3}.}
    \label{fig:iou_coeff}
\end{figure}

\noindent \textbf{Impact of Alignment Criterion Coefficient.} The impact of the alignment coefficient is detailed in Fig. \ref{fig:iou_coeff}, where we evaluate the \textsc{Box Seeker} proposal generation and \textsc{Box Oracle} through different $\alpha_{\text{IoU}} \in \{ 0, 1, 2, 3, 4\}$ under \textcolor{seal}{$\S$ \textsc{Setting} 3}. The alignment criteria improves proposal filtering over density alone, but increasing the coefficient to 2 has a significant benefit, establishing that the 2D alignment provides better guidance for finding novel instances. However, increasing the value further (\textit{e.g.}, 3-4) results in diminished performance as the density can only contribute to ranking by an insignificant amount. 

\begin{figure}[t]
    \centering
    \includegraphics[width=0.8\linewidth]{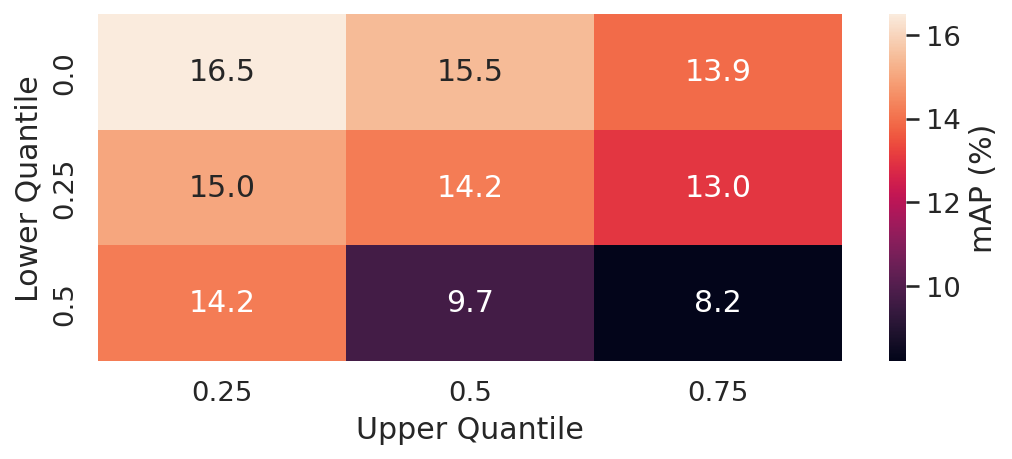}
    \caption{\textbf{Impact of Quantiles with respect to base and novel instance detection under \textcolor{seal}{$\S$ \textsc{Setting} 3}.}\vspace{-2ex}}
    \label{fig:quantile_sensitivity}
\end{figure}

\noindent \textbf{Impact of Search Space Quantiles.} The quantile selection for $d_{\text{min}}$ and $d_{\text{max}}$ is crucial to separating between background and foreground in each frustum. The intuition is to select a large enough lower quantile to exclude any noise before the object, and a small enough upper quantile to remove any background noise. In Fig. \ref{fig:quantile_sensitivity} the lower quantiles $0.0, 0.25, 0.5$ and the upper quantiles $0.25, 0.5, 0.75$ are tested. Clearly, utilising a frustum with $d_{\text{min}}$ and $d_{\text{max}}$ from the quantiles $0.0$ and $0.25$ results in the best performance, as the background is sufficiently removed. Points closer to the camera than the object of interest do not significantly affect the proposal quality, as a lower quantile of $0.0$ results in the best mAP. Overall, increasing the upper quantile results in worse performance as the background is not removed.  

\begin{figure}[H]
\begin{minipage}[t]{.5\linewidth}
    \centering
    \includegraphics[width=\linewidth,height=2.5cm]{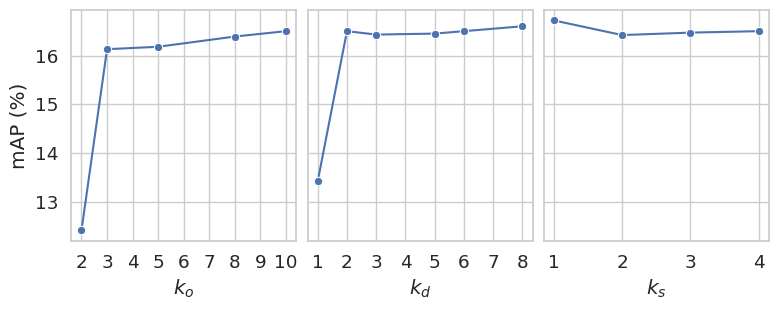} 
    \caption{\textbf{Impact of $k_{\text{o}}, \, k_{\text{d}}, $ and $k_{\text{s}}$.}}
    \label{fig:k_sensitivity}
\end{minipage}
\begin{minipage}[t]{.5\linewidth}
    \includegraphics[width=\linewidth,height=2.5cm]{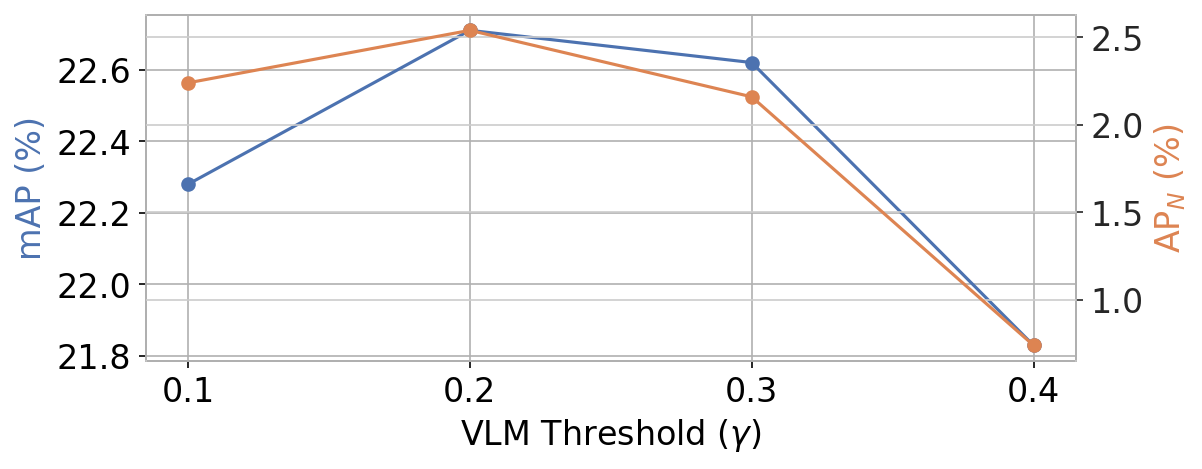}
    \caption{\textbf{Impact of Logit Fusion.} Evaluated with Centerpoint backbone using GLIP and varying $\gamma$.} 
    \label{fig:logit_fusion}
\end{minipage}
\end{figure}

\noindent\textbf{Search Space Sensitivity.} We ablate the \textsc{Greedy Box Seeker} search space in Fig. \ref{fig:k_sensitivity}. We experiment with varying the number of trials for each dimension, including $k_d$ depths, $k_o$ orientations and $k_s$ sizes. We find that we can considerably improve computational efficiency whilst maintaining strong performance. The number of orientations, $k_o$ can be reduced to 3, the number of depths can be reduced to 2 and the number of sizes 1, significantly reducing the number of trials from 320 (maximum) to 6. 

\subsection{Self-Training Objectives}\label{sec:self_training}

In the following section, we provide ablations for our self-training process. We measure the impact of introducing previous work on loss normalisation to mitigate the effect of label noise.

\noindent \textbf{Impact of Self-Training Loss Normalisation.}
We tried to utilise prior work \cite{DBLP:conf/nips/ZophGLCLC020} on self-training to smooth the label noise from novel proposals on Centerpoint. The motivation for the loss is to balance the contributions of the supervised (base classes) and self-supervised (novel classes) during training, by using the exponential moving average (EMA) of the losses to re-weight the self-supervised part. The ratio between the EMA loss on base classes and novel classes is utilised to calculate a coefficient for the novel class loss, with a hyperparameter $\alpha$ to tune the weight of novel classes. Given the loss on base classes $L_B$ and novel classes $L_N$, the loss is calculated as:
\begin{equation*}
    L = \frac{1}{1 + \alpha} (L_B + \alpha \frac{\bar{L}_B}{\bar{L}_N} L_N),
\end{equation*} where $\bar{L}_B$ and $\bar{L}_N$ are exponential moving averages. The loss was applied to the Centerpoint heatmap classification loss. Table \ref{tab:loss_normalisation} displays the results with and without loss normalisation. For both experiments, $\alpha = 0.5$ reduces the impact of noisy proposals by weighting the novel classes as half the base. There is mostly negligible difference in metrics, but loss normalisation gives a small absolute increase in AP$_\text{N}$ of 0.69\%. 

\begin{table}[H]
    \caption{\textbf{Open-vocabulary evaluation with Centerpoint and 4 novel classes (\textcolor{seal}{$\S$ \textsc{Setting} 1}) during training.}}
    \centering
    \begin{tabular}{c S[table-format=2.2]S[table-format=2.2]S[table-format=2.2]S[table-format=2.2]}
    \toprule
Loss Normalisation & \multicolumn{1}{l}{mAP} & \multicolumn{1}{l}{NDS} & \multicolumn{1}{l}{AP$_{\text{B}}$} & \multicolumn{1}{l}{\textcolor{seal}{AP$_{\text{N}}$}}   \\
\midrule
           & 37.38 & \textbf{40.28} & \textbf{49.99} & 18.46 \\
\checkmark & 37.38 & 39.82 & 49.54 & \textbf{19.15} \\
 \bottomrule
    \end{tabular} 
    \label{tab:loss_normalisation}
\end{table}
\subsection{Ablation Study on \textsc{Top-down Projection}}\label{sec:topdown}

\noindent \textbf{Impact of Logit Fusion.} For top-down OV-3D baselines, we test different variants by incorporating the following strategies of logit fusion. During test-time, \textsc{Top-down Projection} methods must decide whether to relabel proposals as unknowns or keep them as the proposed known class. To do this, we set a threshold on the VLM confidence for each 3D proposal, and reject the VLM predicted label if the confidence is not sufficient. The outgoing label for proposal $i$, is given as $\widehat{\mathbf{Y}}_i$, 
\begin{equation}
    \widehat{\mathbf{Y}}_i = \begin{cases}
    \widehat{\mathbf{Y}}^{\operatorname{3D}}_i 
      &   \mathbf{p}^{\operatorname{VLM}}_i \leq \gamma \\
    \widehat{\mathbf{Y}}^{\operatorname{VLM}}_i  & \text{otherwise}\end{cases}.
\end{equation}
As shown in Fig. \ref{fig:logit_fusion}, a VLM confidence threshold of 0.2 provides the best performance balance, and also the highest AP$_\text{N}$.


\subsection{Discussions on CLIP2Scene Zero-shot Results}\label{sec:clip2scene}

\begin{table}
    \centering\caption{\textbf{Zero-shot Lidar Segmentation on nuScenes with CLIP2Scene}. Results are given in \%. Row 1 shows the segmentation results yielded by CLIP2Scene, then rows 2 and 3 show the proposal results with the proposed \textsc{Top-Clustering}. }
    \label{tab:clip2scene}
\resizebox{1.0\linewidth}{!}{
    \begin{tabular}{l  | S[table-format=2.2]S[table-format=2.2]S[table-format=2.2]S[table-format=2.2]S[table-format=2.2]S[table-format=2.2]S[table-format=2.2]S[table-format=2.2]S[table-format=2.2]S[table-format=2.2]}
    \toprule
    Task (Metric) & \multicolumn{1}{l}{Car} &\multicolumn{1}{l}{Const.} &\multicolumn{1}{l}{Trai.} & \multicolumn{1}{l}{Barr.} & \multicolumn{1}{l}{Bic.} & \multicolumn{1}{l}{Ped.} &\multicolumn{1}{l}{Truck} & \multicolumn{1}{l}{Bus} & \multicolumn{1}{l}{Motor.} & \multicolumn{1}{l}{Cone.}  \\
    \midrule
    Zero-shot Segmentation (mIoU) & 29.60 & 9.90 & 0.00 & 7.90 & 1.30 & 0.60 & 45.60 & 43.90 & 9.40 & 18.70 \\
    \midrule
      \textsc{Top-Clustering} \& HDBScan (AP$_\text{3D}$)  & 4.01 & 0.00 & 0.00 & 0.00 & 0.00 & 0.00 & 1.38 & 0.37 & 0.00 & 1.27       \\ 
     \textsc{Top-Clustering} \& DBScan (AP$_\text{3D}$) & 3.07 & 0.00 & 0.00 & 0.00 & 0.00 & 0.00 & 1.86 & 0.72 & 0.00 & 0.00      \\ 
 \bottomrule
    \end{tabular}
}
\end{table}

\begin{figure}[H]
    \centering
    \includegraphics[width=0.6\linewidth]{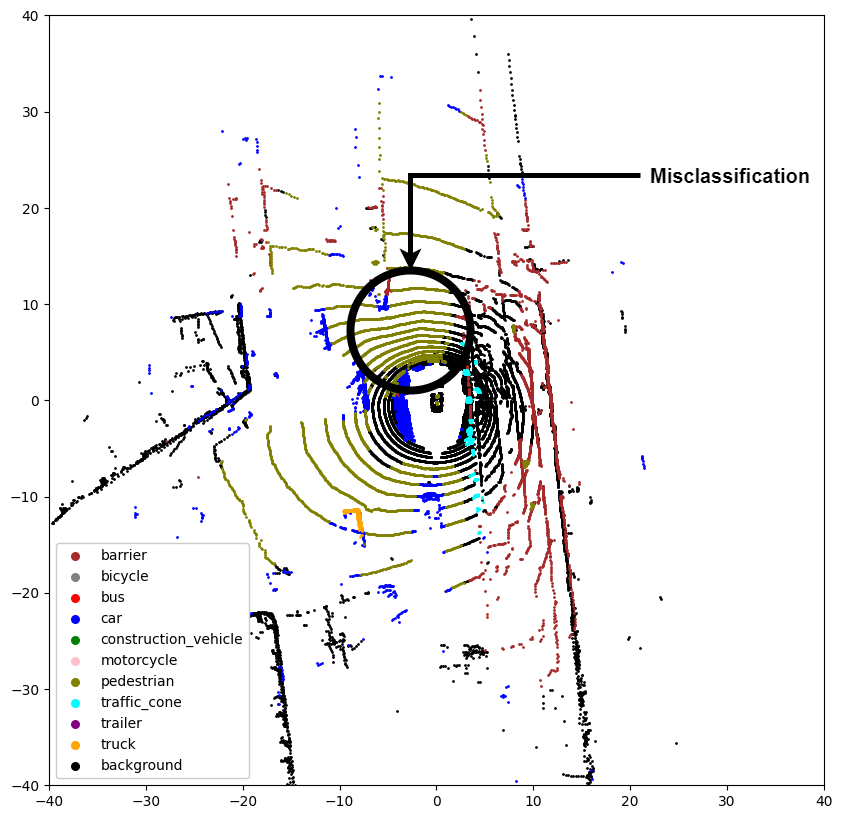}
    \caption{\textbf{CLIP2Scene Zero-shot misclassification.} Many points on the road surface are incorrectly predicted as pedestrian.} 
    \label{fig:clip2scene_misclassification}
\end{figure}

In Table \ref{tab:clip2scene}, we show the zero-shot segmentation results (row 1) and the detection results with \textsc{Top-Clustering} as presented in the main paper. The zero-shot segmentation results (mIoU) are below 10\% for many classes, including \texttt{Construction Vehicle, Trailer, Barrier, Pedestrian} and \texttt{Motorcycle}; for these classes the detection result is \textbf{zero}, likely due to having an \textbf{inferior} pixel-wise classification accuracy, restricting the density of points with the correct label and thus making density clustering unable to find the correct objects. Visualisation of the misclassification error is available in Fig. \ref{fig:clip2scene_misclassification}, where it is clear that the road is associated with pedestrians and therefore segmented incorrectly. Objects with better segmentation accuracy have a likely chance of being correctly partitioned into clusters corresponding to individual object instances, as is the case for the rest of the classes. However, the average precision of these classes is quite low, due to the sparsity of points in a single sweep. 

The lack of AP with this method suggests that the LiDAR scenes in nuScenes are probably too sparse for density-based clustering to work. We utilised 1 sweep for computational efficiency, and due to CLIP2Scene being trained on 1 sweep, further study is possible with LiDAR densification methods (aligning the sweeps) so that it is possible to run these methods with denser point clouds. We compared with 1 sweep to make it fair to our work, which only uses 1 sweep. Multi-sweep necessitates scene flow prediction and tracking, as done in previous works \cite{DBLP:conf/nips/LiPL21, DBLP:conf/cvpr/QiZNSVDA21} however these works consider only vehicles at this stage and need to be expanded to accomodate open vocabularies.

\clearpage  

%
%
\bibliographystyle{splncs04}
\bibliography{main}